%% file: abstraction-alignment.tex
\documentclass[sigconf]{acmart} %

\AtBeginDocument{%
  }

\copyrightyear{2025}
\acmYear{2025}
\setcopyright{cc}
\setcctype{by}
\acmConference[CHI '25]{CHI Conference on Human Factors in Computing Systems}{April 26-May 1, 2025}{Yokohama, Japan}
\acmBooktitle{CHI Conference on Human Factors in Computing Systems (CHI '25), April 26-May 1, 2025, Yokohama, Japan}
\acmDOI{10.1145/3706598.3713406}
\acmISBN{979-8-4007-1394-1/25/04}

\input{sections/00_includes}
\input{sections/00_macros}

\begin{document}

\title[Abstraction Alignment]{Abstraction Alignment: Comparing Model-Learned and Human-Encoded Conceptual Relationships}

\author{Angie Boggust}
\affiliation{%
  \institution{CSAIL\\Massachusetts Institute of Technology}
  \city{Cambridge}
  \state{Massachusetts}
  \country{USA}
}
\email{aboggust@mit.edu}
\orcid{1234-5678-9012}

\author{Hyemin Bang}
\affiliation{%
  \institution{CSAIL\\Massachusetts Institute of Technology}
  \city{Cambridge}
  \state{Massachusetts}
  \country{USA}
}
\email{hbang@mit.edu}

\author{Hendrik Strobelt}
\affiliation{%
  \institution{IBM Research AI}
  \city{Cambridge}
  \state{Massachusetts}
  \country{USA}
}
\email{hendrik@strobelt.com}

\author{Arvind Satyanarayan}
\affiliation{%
  \institution{CSAIL\\Massachusetts Institute of Technology}
  \city{Cambridge}
  \state{Massachusetts}
  \country{USA}
}
\email{arvindsatya@mit.edu}

\renewcommand{\shortauthors}{Boggust et al.}

\begin{abstract}

\input{sections/00_abstract}

\end{abstract}



\input{figures/teaser}

\maketitle

\input{sections/01_introduction}
\input{sections/02_related_work}

\input{sections/03_method}

\input{sections/04_case_studies}
\input{sections/05_discussion}

\input{sections/06_conclusion}

\begin{acks}
We thank David Bau, Been Kim, and Martin Wattenberg for their feedback, which significantly improved this work.
We also thank our study participants for sharing their experiences and insights.  
This work was supported by NSF grant \#1900991. 
The first author is supported by the Apple Scholars in AIML PhD Fellowship.
\end{acks}

\bibliographystyle{ACM-Reference-Format}
\bibliography{abstraction_alignment}

\newpage
\appendix
\input{sections/appendix}

\end{document}

%% file: sections/00_includes.tex
\usepackage[capitalise]{cleveref}
\crefname{figure}{Figure}{Figures}
\usepackage{xspace}
\usepackage{enumitem}
\usepackage{amsmath}

\usepackage{listings}
\usepackage{xcolor}
\usepackage{float}

%% file: sections/00_macros.tex
\newcommand{\concept}[1]{\textsc{#1}}
\newcommand{\inlinequote}[1]{``\textit{#1}''}
\newcommand{\function}[1]{\texttt{#1}}
\newcommand{\abstraction}[1]{---\textit{#1}$\rightarrow$}
\newcommand{\query}[1]{\texttt{#1}}

\newcommand{\revision}[1]{#1}

\newcommand{\codeURL}{\url{https://github.com/mitvis/abstraction-alignment}\xspace}
\newcommand{\interfaceURL}{\url{https://vis.mit.edu/abstraction-alignment/}\xspace}

\newcommand{\sref}[1]{\S\ref{#1}}

\Crefname{lstlisting}{Listing}{Listings}
\definecolor{lightgray}{HTML}{F7F7F7}
\definecolor{Periwinkle}{HTML}{7977B8}
\definecolor{Thistle}{HTML}{D883B7}

\newcommand{\abstractionfit}{{\function{abstraction} \function{match}}\xspace}
\newcommand{\subgraphpreference}{{\function{subgraph}\:\function{preference}}\xspace}
\newcommand{\conceptcoconfusion}{{\function{concept} \function{co-confusion}}\xspace}

%% file: sections/00_abstract.tex
While interpretability methods identify a model’s learned concepts, they overlook the relationships between concepts that make up its abstractions and inform its ability to generalize to new data.
To assess whether models’ have learned human-aligned abstractions, we introduce abstraction alignment, a methodology to compare model behavior against \revision{formal} human knowledge.
Abstraction alignment externalizes domain-specific human knowledge as an abstraction graph, a set of pertinent concepts spanning levels of abstraction. 
Using the abstraction graph as a ground truth, abstraction alignment measures the alignment of a model’s behavior by determining how much of its uncertainty is accounted for by the human abstractions. 
By aggregating abstraction alignment across entire datasets, users can test alignment hypotheses, such as which human concepts the model has learned and where misalignments recur. 
In evaluations with experts, abstraction alignment differentiates seemingly similar errors, improves the verbosity of existing model-quality metrics, and uncovers improvements to current human abstractions.

%% file: figures/teaser.tex
\begin{teaserfigure}
  \includegraphics[width=\textwidth]{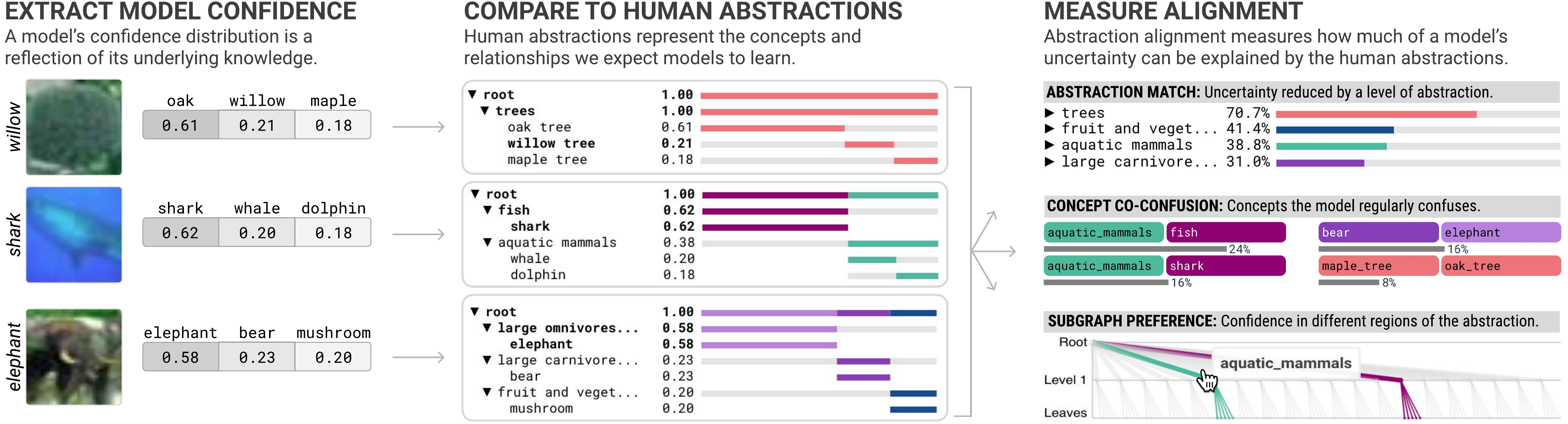}
  \caption{
  Abstraction alignment measures human-AI alignment by comparing model behavior to known human abstractions.
  }
  \Description{A depiction of the abstraction alignment methodology. On the left "Extract Model Confidence" shows three images and their probability distributions. In the middle "Compare to Human Abstractions" shows the probability distributions for each image propagated in a human abstraction graph. Finally "Measure Alignment" shows three metrics: "abstraction match", "concept co-confusion", and "subgraph preference".}
  \label{fig:teaser}
\end{teaserfigure}

%% file: sections/01_introduction.tex
\section{Introduction}
AI alignment is increasingly critical to meet growing societal and regulatory demands for AI systems that make human-like decisions~\citep{terry2023ai, sucholutsky2023getting}.
To meet these demands, the research community has developed interpretability methods to uncover the concepts models use to reason about their inputs and generate outputs\,---\,for instance, using \concept{wheels} to classify car images~\cite{ghorbani2019automating, olah2020zoom} or \concept{tourist attractions} to respond to travel text~\cite{templeton2024scaling}.
By analyzing these concepts, users can better understand model behavior and identify potential misalignments, such as an overreliance on one medical concept when making complex diagnoses~\citep{kim2018interpretability, boggust2022shared} or a propensity for sycophantic responses at the expense of truthfulness~\citep{templeton2024scaling}.

However, existing methods analyze a model's learned concepts in isolation, quantifying the model's sensitivity to each concept independently~\citep{kim2018tcav,templeton2024scaling}.
While such testing procedures identify the importance of each concept to the model's decision, they overlook the model's learned relationships between concepts.
Yet, conceptual relationships are core to the ability to extrapolate learned concepts to new data, contextualize knowledge across tasks, and flexibly reason at a task-appropriate level of specificity~\citep{yee2019abstraction,alexander2018pattern,liskov1986abstraction,eva2005every}. 
Interpreting these characteristics of a model requires analyzing its \textit{abstractions}, validating that it has not only learned granular concepts (e.g., \concept{schnauzer}), but that it also organizes them into progressively more general notions (e.g., \concept{dog} and then \concept{animal}).

Although existing interpretability techniques have been used to discover that models learn concepts at varying levels of detail~\citep{kim2018interpretability, ghorbani2019automating, hernandez2021natural, bau2017network, oikarinen2022clip}, analyzing abstractions requires significant manual effort.  
Users must survey interpretability results piece-by-piece to confirm that the model's abstractions are human-aligned, checking that all concepts activate as expected~\citep{kim2018interpretability, templeton2024scaling}, the set of concepts is comprehensive for the task~\citep{templeton2024scaling}, and similar concepts are represented similarly by the model~\citep{templeton2024scaling,bricken2023towards}.
In each of these cases, the process of estimating alignment occurs largely inside the user's head, requiring significant cognitive effort to compare model concepts against their domain abstractions.
Moreover, by relying on individuals' mental abstractions, existing approaches limit alignment assessment to users with deep domain-specific knowledge, such as medical specialists~\citep{kim2018interpretability, boggust2022shared} or chess Grandmasters~\citep{schut2023bridging}.

To scaffold the process of assessing model alignment, we introduce \emph{abstraction alignment}, a methodology to measure the agreement between a model's learned abstractions and an explicit human representation of the modeled domain.
Abstraction alignment is a form of representational alignment~\citep{sucholutsky2023getting} that compares model outputs (a proxy for its internal representations) against codified human abstractions (a proxy for our internal representations).
Abstraction alignment begins with a \emph{human abstraction graph}\,---\,an agreed upon representation of \revision{formal} human knowledge containing a set of pertinent concepts spanning levels of abstraction, such as a lexical graph~\citep{miller1995wordnet} or medical hierarchy~\citep{worldhealthorganization1978icd9}.
It then measures alignment by evaluating how well the abstraction graph accounts for a model's decision uncertainty.
Through this process, model output probabilities (i.e., confidence in each class or token) are mapped to concepts in the human abstraction graph, and the probabilities of sibling concepts (e.g., \concept{guitarist} and \concept{singer}) are summed and propagated to shared ancestors (e.g., \concept{musician} and \concept{artist}). 
The result is the model's \textit{fitted abstraction graph}, representing its decision making process across concepts at many levels of abstraction.

To aggregate abstraction alignment over many model decisions, we define three metrics.
\texttt{Abstraction match} measures how much of the model's confusion is mitigated by moving up a level of abstraction, \conceptcoconfusion tests how often the model confuses concepts, and \subgraphpreference quantifies which abstractions the model prefers.
By integrating the abstraction alignment metrics into an interactive interface, we enable users, ranging from computer scientists to medical domain experts, to ask and answer alignment hypotheses, such as which human concepts the model has learned and what recurring misalignments the model makes.

We demonstrate how abstraction alignment facilitates alignment analysis through case studies with expert users interpreting an image classification model, benchmarking generative language models, and auditing a clinical ML dataset.
In an image classification task, abstraction alignment helps interpret model behavior by distinguishing problematic misalignments from benign low-level errors.
Analyzing abstraction alignment across the entire test dataset reveals that most model mistakes are not abstraction-aligned, as it learns to differentiate images using visual abstractions like \concept{color} instead of the desired human abstractions based on \concept{biological} and \concept{usage} differences. 
These misalignments suggest instances where the model may fail, ways to improve data collection, and possible recategorizations of the human abstraction graph.

Then, we collaborate with three language model researchers, applying abstraction alignment to their alignment task: investigating the specificity of generative language models.
While users currently test model specificity by comparing the model's generated text against a few synonymous words, abstraction alignment expands the verbosity of these benchmarks by testing thousands of words across numerous levels of abstraction.
As such, we find that abstraction alignment allows researchers to test more complex hypotheses, such as the model's preferred level of specificity and which unrelated words the model commonly confuses.

Finally, in a participatory AI setting, four medical experts analyze abstraction alignment to audit the MIMIC-III clinical dataset~\citep{johnson2016mimica,johnson2016mimicb}, testing whether its labels reflect appropriate medical abstractions.
With abstraction alignment, experts uncover discrepancies between the medical abstractions we expect models to learn and those codified in the dataset.
For instance, experts find that how diseases are classified in the dataset does not always align with the World Health Organization's (WHO) standards~\citep{worldhealthorganization1978icd9}.
These abstraction misalignments suggest data processing strategies that better reflect human expectations and exposes known issues in the disease hierarchy, some of which have been recently addressed by the WHO~\citep{cartwright2013icd}.
These results signal that, beyond improving the alignment of model representations, abstraction alignment may also inspire improvements to existing human representations.

Abstraction alignment is publicly available with open-source code at \codeURL and an interactive interface at \interfaceURL.

%% file: sections/02_related_work.tex
\section{Related Work}
\subsection{Abstraction and Human Knowledge}
Abstraction is the process of distilling many individual data instances into a set of fundamental concepts and relationships that capture essential characteristics of the data~\citep{yee2019abstraction, alexander2018pattern, liskov1986abstraction}.
It is a key feature of human cognition as it allows us to flexibly reason at the level of specificity appropriate for our task and generalize our knowledge by fitting abstracted patterns to new data~\citep{yee2019abstraction, victor2011up,eva2005every}.
As a result, abstractions form the basis for human information encodings across domains like linguistics~\citep{miller1995wordnet, dewey2011dewey}, biology~\citep{hinchliff2014synthesis, linnaeus1758systema}, and medicine~\citep{worldhealthorganization1978icd9,who2022icd10}.
In machine learning, abstractions are built into many tasks, including image classification~\citep{krizhevsky2009learning, deng2009imagenet} and medical coding~\citep{johnson2016mimica, johnson2016mimicb}.
Even datasets without built-in abstractions are often linked to existing abstractions by matching their outputs to corresponding concepts~\citep{redmon2017yolo9000}.
Encouragingly, researchers have recently integrated human abstractions into model training pipelines, resulting in increased model generalization~\citep{muttenthaler2024aligning}, and advocated for using conceptual relationships, like abstractions, to advance our understanding of foundation models~\citep{wattenberg2024relational}.
Building on this rich history, abstraction alignment leverages abstractions to better understand human-AI alignment.

Related research has studied formal representations of human knowledge, known as knowledge graphs~\citep{hogan2021knowledge,ji2020survvey}.
Knowledge graphs reflect the relationships between entities, like distance (\concept{Eiffel Tower} \abstraction{is near} \concept{Arc de Triomphe}) or connectivity (\concept{BOS} \abstraction{direct flight} \concept{MEX})~\citep{hogan2021knowledge}.
In abstraction alignment, we represent human abstractions as an abstraction graph, a type of knowledge graph where nodes are concepts and edges encode abstractions from specific to general concepts (e.g., \concept{cardiologist} \abstraction{type of} \concept{doctor} or \concept{Montreal} \abstraction{located in} \concept{Quebec}).
We make this distinction because we are interested in understanding whether a model has learned to reason with human-like abstractions.
These human abstraction graphs provide an explicit representation of \revision{formal} human knowledge (\cref{sec:method-human-abstraction}) that allows us to quantify alignment.

\subsection{AI Alignment and Interpretability}
Aligned with our goal of understanding machine learning models, interpretability research measures model reliance on known human concepts~\citep{doshivelez2017rigorous, rai2020explainable}.
For instance, saliency methods highlight relevant input features~\citep{selvaraju2017grad, carter2019made, boggust2022shared, smilkov2017smoothgrad, petsiuk2018rise};
example-based methods derive influential inputs~\citep{zeng2021human,wu2019errudite,mothilal2020explainaing,koh2017understanding};
feature visualizations identify concepts that activate model neurons~\citep{olah2017feature, erhan2009visualizing, bau2017network,hernandez2021natural,li2021implicit};
and concept-based~\citep{kim2018tcav, ghorbani2019automating} and mechanistic methods~\citep{hernandez2023measuring,li2021implicit,li2022emergent,bricken2023towards,merullo2023circuit,olah2020zoom,elhage2022toy,templeton2024scaling} identify human concepts encoded in a model's latent space.
Together these methods have identified problematic model correlations~\citep{carter2021overinterpretation}, made sense of complex neuron activations~\citep{olah2018building, carter2019activation,li2022emergent,nanda2024actually}, and discovered novel concepts that advance human understanding~\citep{schut2023bridging}.
Building on their success, abstraction alignment expands these methods from identifying independent concepts to understanding the relationships between them, ensuring that models learn human-aligned concepts and abstractions.

At the same time, visualization research has explored how to communicate interpretability results to users.
Visualizing interpretability results\,---\,e.g., saliency heatmaps~\citep{Kapishnikov2019xrai,smilkov2017smoothgrad,Boggust2022Saliency,Schuff2022Human}, embedding scatterplots~\citep{boggust2022embedding,sivaraman2022emblaze,wattenberg2016how}, and feature dictionaries~\citep{carter2019exploring,templeton2024scaling,lin2023neuronpedia}\,---\,has enabled greater meaning-making from experts and engagement from lay users~\citep{bhat2024designing}.
Interactive interfaces~\citep{bauerle2022symphony,cabrera23zeno,robertson2023angler,strobelt2018seq,wexler2019what} have allowed users to perform error analysis~\citep{wu2019errudite, suresh2023kaleidoscope}, track model provenance~\citep{boggust2024compress,tensorflow}, inspect decision boundaries~\citep{sohns2023decision}, and intervene on models~\citep{hohman2024talaria}.
We instantiate abstraction alignment in an interactive interface (\cref{sec:interface}), allowing users, from ML researchers to domain experts, to actively participate in alignment tasks.

Abstraction alignment also follows a rich history of human-AI alignment research~\citep{terry2023ai,sucholutsky2023getting}, studying methods for measuring~\citep{Langlois2021Passive,bang2024explanation,Muttenthaler2023human,Oktar2023Dimensions,rane2024concept}, bridging~\citep{Gupta2017Aligned,Ramesh2022Hierarchical,schut2023bridging}, and increasing~\citep{Hinton2015Distilling,Tian2019Contrastive,muttenthaler2024aligning,Phuong2019Towards} the alignment of model and human representations.
Within the alignment framework developed by \citet{sucholutsky2023getting}, abstraction alignment is a form of behavioral alignment~\citep{Geirhos2020Beyond,Ouyang2022Training} where model outputs serve as proxies for its internal representations.
Our experiments examine representational alignment by comparing model representations (proxied by model outputs) and human representations (proxied by the human abstraction graph) across a set of evaluation data.
In \cref{sec:method-formalism}, we characterize abstraction alignment using the alignment framework developed by \citet{sucholutsky2023getting}.
Unlike prior studies that focus on a single metric to quantify and optimize alignment, abstraction alignment offers a methodology for evaluating how closely model behaviors reflect \revision{formal} human knowledge.
As a result, there are many possible metrics that capture specific aspects of abstraction alignment (we define three in \cref{sec:method-metrics}) and abstraction alignment facilitates qualitative, interactive human analysis (\cref{sec:case-studies}).

\subsection{Uncertainty Estimation}
Abstraction alignment relies on model uncertainty to compute alignment, working under the assumption that the model's uncertainty reflects its learned abstractions.
Model uncertainty can be broadly categorized as aleatoric~\citep{northcutt2021pervasive,VillaNovaRodrigues2021information}, arising from irreducible observational noise like noisy data and labeling errors, or epistemic~\citep{Wang2024Epistemic}, stemming from limited knowledge like insufficient training data or out-of-distribution inputs~\citep{hullermeier2021aleatoric,fox2011distinguishing,kiureghian2009aleatoric,zerva2022disentangling}.
Relatedly, uncertainty quantification research focuses on accurately extracting these uncertainties from models such that the model's purported confidence is an interpretable measure of correctness~\citep{abdar2021review,gal2016dropout,lakshminarayanan2017simple,guo2017on,kumar2019verified,zadrozny2001obtaining,Chen2022Interpretable}.
Instead of classifying or adjusting model uncertainty, abstraction alignment uses it as a proxy for the model's internal representations.
As a result, abstraction alignment is agnostic to the type of uncertainty, because both types reflect the model's internal conceptual boundaries and future behavior.
Whether the uncertainty arises because the model lacks training data to distinguish a concept (epistemic) or because humans also confuse the concept (aleatoric), it nevertheless represents the model’s understanding of that concept.

%% file: sections/03_method.tex
\section{The Abstraction Alignment Methodology}
\label{sec:method}

The goal of abstraction alignment is to measure how well a model's behavior aligns with human abstractions.
Our method is based on the assumption that a model's uncertainty is a reflection of its learned abstractions.
That is, concepts the model commonly confuses are more similar in its abstractions than concepts it perfectly separates.
For instance, if a model has learned human abstractions, then, in aggregate, it should be more likely to confuse \concept{apples} with other \concept{fruits} than with unrelated concepts, like \concept{motorcycles}.

While there are likely many methods for measuring abstraction alignment, we take a post hoc and model-agnostic approach that compares model outputs against existing human abstractions.
To compute abstraction alignment, we represent human abstractions as an \textit{abstraction graph} (\cref{sec:method-human-abstraction}).
We compare the model's behavior to human abstractions by mapping the model's output options (e.g., its classes or tokens) to concept nodes in the human abstraction graph (\cref{sec:method-model-outputs}).
Given a dataset instance, like an image or a sentence, we compute the model's \textit{fitted abstraction graph}, a weighted version of the abstraction graph representing the model's confidence in a range of concepts across multiple levels of abstraction.
We use the model's fitted abstraction graphs to define abstraction alignment metrics that quantify how well the human abstractions account for the model's behavior (\cref{sec:method-metrics}).

\input{code/abstraction_alignment_propagation}

\subsection{Representing Human Abstractions}
\label{sec:method-human-abstraction}
Abstraction alignment shifts the alignment process from mentally estimating alignment using a human's internal knowledge to externally inspecting precomputed alignment results.
To do so, we externalize \revision{formal} human knowledge as a directed acyclic graph (DAG) called the \textit{human abstraction graph}.
We define an abstraction graph as a type of knowledge graph
where nodes represent concepts and edges represent abstraction relationships between precise and broad concepts.
For example, in the medical abstraction graph in \cref{sec:medical-dataset-analysis}, nodes represent medical diagnoses and edges represent the abstractions between specific diagnoses, like \concept{frontal sinusitis}, and broader diagnostic categories, like \concept{respiratory infections}~\citep{worldhealthorganization1978icd9}.
Formally, the human abstraction graph is a DAG $G$ containing a set of nodes $N := \{n_k\}$.
Nodes are distributed across levels $L:=\{l_h\}$ where each $l_h \subseteq N$, and a node's level is defined as the length of the longest path ($h$) from the node to a root.

DAGs are well suited to representing human abstractions because they efficiently encode both human concepts and abstraction relationships.
We can easily access a concept's level of abstraction by measuring its height and move up and down levels of abstraction by accessing its ancestors or descendants.
Since the graph is acyclic, it guarantees the hierarchical structure that underpins abstraction.
Further, DAGs are commonly used to represent human abstractions~\citep{worldhealthorganization1978icd9, miller1995wordnet} and are built into ML datasets~\citep{krizhevsky2009learning, deng2009imagenet, johnson2016mimica, johnson2016mimicb}, allowing abstraction alignment to apply to various domains.

We use human abstraction graphs as agreed-upon, external representations of \revision{formal} human knowledge.
While they may not perfectly match any individual's internal abstractions, they are useful proxies as they reflect collective human meaning-making.
For example, while we may not individually know every word and relation in the WordNet lexical graph, it nevertheless represents collective English language abstractions that we use to communicate~\citep{miller1995wordnet}.
These graphs are often shared between individuals~\citep{Tudorache2008supporting,Sure2002ontoedit,noy2001ontology,musen1992dimensions}, used to educate newcomers to a domain~\citep{worldhealthorganization1978icd9}, and employed when building additional knowledge representations~\citep{krizhevsky2009learning, deng2009imagenet}.

\subsection{Integrating Model Outputs with Human Abstractions}
\label{sec:method-model-outputs}
To compare the model's behavior against human abstractions, we map the model's output space (e.g., its classifiable classes or generatable tokens) to nodes in the human abstraction graph.
This defines a mapping from each of the model's outputs $O:=\{o_j\}$ to a node $n_k \in G$ that corresponds to the same concept.
Often this mapping is straightforward because the human abstraction graph is built into the modeling task\,---\,e.g., the CIFAR-100 classes are mapped to higher-level concepts~\citep{krizhevsky2009learning} (\cref{sec:interpreting-cifar-models}).
However, even when the human abstraction graph is separate from the modeling task, the model's outputs can often easily be mapped to the graph's nodes.
For instance, in \cref{sec:language-model-specificity}, we map language model tokens to words in the WordNet lexical graph~\citep{miller1995wordnet} by searching WordNet for the token's most similar definition.

With a mapping from model output to concept node, we can now compare the model's behavior against the human abstractions.
To do so, we compute a \textit{fitted abstraction graph}, a weighted version of the human abstraction graph representing the model's confidence in each concept for a given decision.
Following the algorithm in \cref{lst:abstraction-fit}, we compute a fitted abstraction graph for every instance in an evaluation dataset $D := \{d_i\}$.
Given an instance $d_i$, like an image or sentence, we extract the model's probability for each possible output concept $o_j$.
Then, we assign each node in the human abstraction graph $n_k$ a value $v_{ik}$ equal to the model's probability for the node's concept or any of its descendants.
For example, given a CIFAR-100 image classification model as in \cref{sec:interpreting-cifar-models}, the value of \concept{flower} is the sum of the model's confidence that the given image is a \concept{poppy}, \concept{rose}, \concept{tulip}, \concept{orchid}, or \concept{sunflower}.
By propagating the model's probabilities through the abstraction graph, the fitted abstraction graph provides a measure of the model's confidence in a range of concepts across levels of abstraction.

So far, we have defined fitted abstraction graphs using an ML model, but we can also use them to represent other types of encoded abstractions.
We can compute a fitted abstraction graph for any function that maps dataset instances $d_i$ to a distribution over human concepts $O$.
This function, $f: D\mapsto\mathbb{R}^{|O|}$, is often the forward function of a machine learning model, such as an image classifier (\cref{sec:interpreting-cifar-models}) or language generation model (\cref{sec:language-model-specificity}).
However, as we demonstrate in \cref{sec:medical-dataset-analysis}, this function can also represent the information encoded in a dataset, where $f$ maps clinical notes $d_i$ to clinical codes $o_j$, assigning each $o_j$ a value based on whether a human labeled the note with that code.
In this case, the fitted abstraction graphs represent the alignment between human labeling patterns $f$ and expected medical abstractions $G$.

\subsection{Measuring Abstraction Alignment}
\label{sec:method-metrics}
The model's fitted abstraction graphs support various alignment hypotheses, such as identifying concepts prone to misalignment and determining the model's preferred level of abstraction.
While there are many alignment metrics one could define across the fitted abstraction graphs, we propose three metrics that have proven useful in our alignment analysis of computer vision classification models, generative language models, and medical datasets (\cref{sec:case-studies}).

\paragraph{Abstraction Match}
One way to measure abstraction alignment is to measure how well the human abstractions account for the model's uncertainty.
If the model's confusion is substantially reduced by moving up a level of abstraction, then the model's behavior is more abstraction-aligned than if it continues to be confused at higher-levels of abstraction.
While there are cases when the model's confusion may acceptably not fit the human abstractions, such as confusion on an image containing multiple objects, in aggregate we expect the model's uncertainty to reflect its abstractions\,---\,i.e., it will confuse concepts that it considers similar.

We measure \abstractionfit as the amount of the model's decision entropy that is reduced by moving up a level of the abstraction graph.
Given two levels ($l_g$ and $l_h$), we compute the difference in entropy, $H(V) = -\sum_{v\in V}v\log(v)$ \citep{shannon1948mathematical}, between the node values $V:=\{v_{ik}\}$ at each level.
The larger the entropy, the more confused the model is across concepts at that level of abstraction.
If the entropy decreases substantially then the model's behavior aligns with the abstraction mapping the low-level nodes to the higher-level nodes.
We aggregate \abstractionfit across a set of data instances $D$, which can be a single instance, the entire dataset, or an informative data subset.
{\small
\begin{equation}
    M(l_g, l_h) = \frac{1}{|D|}\sum_{i=1}^{|D|} H([v_{ik}\, \forall \, n_k \in l_h]) - H([v_{ik}\, \forall \, n_k \in l_g])\label{eq:abstaction-fit}
\end{equation}
}

\paragraph{Subgraph Preference}
Another valuable metric is to compare the values of different fitted abstraction subgraphs.
For instance, in \cref{sec:language-model-specificity}, we compare subgraphs that represent different concepts (e.g., any location vs. \concept{canadian} locations) and different levels of abstraction (e.g., concepts more specific than \concept{journalist} to concepts more general than \concept{journalist}). 
In aggregate, these comparisons help us quantify and compare abstractions the model prefers.

We compute \subgraphpreference by measuring how often the aggregate value of a node in one subgraph, $s_a$, is larger than the aggregate value of a node in another subgraph $s_b$.
This is an extension of the specificity testing metric proposed by \citet{huang2023can}, where $s_a$ represents the specific concept and $s_b$ represents the general concept.
However, while the prior metric was designed to test two concepts, abstraction alignment allows us to test a breadth of concepts, including different levels of abstraction, multiple similar concepts, and concepts related to different abstractions.
If the model's outputs span multiple levels and many concepts in the abstraction graph, we can either compute \subgraphpreference using the nodes' values (as in \cref{sec:language-model-specificity-qualitative}) or use the unpropagated model probabilities as the nodes' values (as in \cref{sec:language-model-specificity-quantitative}).
{\small{
\begin{equation}
    P(s_a, s_b) = \frac{1}{|D|} \sum_{i=1}^{|D|} \mathbf{1}[\max([v_{ik}\, \forall\, n_k \in s_a]) > \max([v_{ik}\, \forall\, n_k \in s_b])] \label{eq:subgraph-preference}
\end{equation}
}}

\paragraph{\revision{Concept Co-confusion}}
Finally, the \conceptcoconfusion metric allows us to measure how often a model assigns probability to pairs of concepts.
While concepts that are ancestors or descendants of each other will definitionally have high \conceptcoconfusion, unrelated concepts with high \conceptcoconfusion are unrelated human concepts that the model deems similar.

To compute \conceptcoconfusion for a pair of nodes, we compute the entropy ($H$)~\citep{shannon1948mathematical} of their values divided by the maximum possible entropy for a pair of nodes.
By computing the entropy, we weight the \conceptcoconfusion by how confused the two nodes are\,---\,e.g., \conceptcoconfusion for nodes with values 0.4 and 0.6 will be higher than nodes with values 0.9 and 0.1 because the model is more confused between the first pair of concepts.
We compute \conceptcoconfusion over the data $D$ to identify repeated confusion.
\begin{align}
    C(n_k, n_l) =   \frac{\sum_{i=1}^{|D|} H([v_{ik}, v_{il}])}{\log_e(2)} \label{eq:concept-coconfusion}
\end{align}

\input{figures/interface_figure}

\subsection{Formalizing Abstraction Alignment as Representational Alignment}
\label{sec:method-formalism}

Representational alignment is a paradigm for measuring the similarity of two systems' internal representations~\citep{sucholutsky2023getting}.
Here, we define abstraction alignment using the representational alignment formalism from \citet{sucholutsky2023getting}.
Abstraction alignment compares machine learning model or dataset abstractions (system $A$) against \revision{formal} human knowledge (system $B$) across a set of dataset instances ($D$).
We use the model's decisions ($Q$) as a proxy for its internal representations and a human abstraction graph ($W$) as a proxy for internal human representations.
We compute abstraction alignment by comparing $Q$ and $W$ using the fitted abstraction graphs (\cref{sec:method-model-outputs}) and abstraction alignment metrics (\cref{sec:method-metrics}).

\paragraph{Data $D$:}
We measure abstraction alignment across a set of evaluation data $D := \{d_i\}$. 
In our experiments, $D$ consists of images or text, but abstraction alignment applies to any data modality.

\paragraph{System $A$:}

System $A$ refers to either the machine learning model or dataset under investigation. When $A$ is a model, the focus is on measuring the alignment of its behavior. When $A$ is a dataset, the goal is to evaluate the alignment of its labels. 
Accordingly, $f: D \mapsto \mathbb{R}^{|O|}$ represents either the model's function mapping inputs to its probability distribution over outputs or the function mapping dataset instances to their labels.

\begin{itemize}
    \item \textit{Measurements $X$:} 
    System $A$'s measurements, $X \in \mathbb{R}^{|D|\times |O|}$, is a matrix of model outputs or dataset labels obtained by applying $f$ to each data instance, $X:=[f(d_1),...,f(d_{|D|})]$.
    
    \item \textit{Embeddings $Q$:} 
    Since abstraction alignment directly studies the model outputs or dataset labels, we let $Q:=X$.
\end{itemize}

\paragraph{System $B$:}

System $B$ represents \revision{formal} human knowledge. 
Accordingly, $g : D \mapsto \mathbb{R}^{|G|}$ maps the entire dataset $D$ to a relevant human abstraction graph $G$ containing nodes $N$. 
It synthesizes domain-specific human knowledge (represented by the data) into core concepts and their relationships (represented by the graph).
\begin{itemize}
        \item \textit{Measurements $Y$:} 
        System $B$'s measurements, $Y \in \mathbb{R}^{|G|}$, are a human abstraction graph relevant to the data, $Y := g(D) = G$.
        For comparison with system $A$, we assume that the output concepts are a subset of the human concepts $O \subseteq N$ in $G$.
        
        \item \textit{Embeddings $W$:} Since $Y$ already represents human abstractions, we do not apply additional transformations ($W:=Y$).
\end{itemize}

\paragraph{Alignment Function $\delta(Q,W)$:}

To compare the model outputs or dataset labels ($Q$) against the human abstraction graph ($W$), we first create a fitted abstraction graph by projecting the concepts and probabilities in $Q$ onto the graph of concepts in $W$ (\cref{sec:method-model-outputs}).
Instead of providing a single quantifier of alignment, this graph-based comparison allows us to define a family of metrics, measuring multiple facets of alignment (\cref{sec:method-metrics}).

\section{The Abstraction Alignment Interface}
\label{sec:interface}

\input{figures/interactivity_figure}

We instantiate abstraction alignment in an interactive visual interface (\cref{fig:interface}) consisting of six interconnected components that express the abstraction alignment metrics (\cref{sec:method-metrics}) and fitted abstraction graphs (\cref{sec:method-model-outputs}).
Users can interactively test alignment hypotheses by filtering to individual concepts, selecting entire graph regions, and querying for alignment patterns (\cref{fig:interface-interactivity}).

\paragraph{Cumulative Fitted Abstraction Graph}
The cumulative fitted abstraction graph (\cref{fig:interface}A) serves as an overview of the model's fitted abstraction graphs and a visualization of the human abstraction graph.
We compute it by summing the fitted abstraction graph (\cref{sec:method-model-outputs}) for every dataset instance such that a node's value in the cumulative fitted abstraction graph is the sum of that node's value across every model decision.
We visualize it as a vertical graph across the levels of abstraction, from the most abstract (root) to the most specific concepts (leaves).
To display meaningful visual groupings, we color nodes and edges based on their level-1 ancestor.
Node radius and edge width reflect the cumulative value assigned to each concept.
For instance, in \cref{fig:interface}A, the thick red edge for \concept{professional} indicates that is assigned high confidence in nearly every model prediction.
To minimize overlapping edges, we use a recursive depth-first layout and sort children based on their value.
Hovering over a node reveals its name, value, and contributing instances, clicking selects the concept and its relatives, and double-clicking selects just that concept (\cref{fig:interface-interactivity}A).
Upon selection, the interface updates to display the relevant dataset instances.

\paragraph{Abstraction Match}
The abstraction match component (\cref{fig:interface}B) instantiates the \abstractionfit metric (\cref{eq:abstaction-fit}), highlighting abstractions the model has learned most effectively.
For every concept and level of abstraction, we compute the proportional \abstractionfit between that level and the next using every dataset instance whose label is a descendant of the concept.
For example, for occupation prediction, the level-1 \abstractionfit for \concept{scientist} is shown as the percent decrease in entropy by moving from level-2 to level-1 over all instances whose true occupation is a descendant of \concept{scientist} (e.g., \concept{physicist} or \concept{astronomer}).
The abstraction match component is displayed as a nested horizontal bar chart, with bars corresponding to concepts in the human abstraction graph and their lengths representing their percent \abstractionfit. 
Selecting a bar (\cref{fig:interface-interactivity}B) updates the interface, displaying results for the set of instances that contributed to that concept's \abstractionfit score.

\paragraph{Concept Distribution}
The concept distribution component visualizes how dataset labels are distributed across levels of abstraction.
For each concept in the human abstraction graph, its concept distribution value represents the number of dataset instances whose label is a descendant of that concept.
For example, in \cref{fig:interface}C, the concept distribution component shows that 1,238 instances in the occupation prediction dataset are labeled as a type of \concept{communicator}.
Like abstraction match, the concept distribution is displayed as a nested horizontal bar chart.
Selecting a bar updates the interface to display dataset instances labeled under that concept (\cref{fig:interface-interactivity}C).

\paragraph{Query}
The query component allows users to search for types of model behavior defined over the abstraction graph (\cref{fig:interface}D).
We define an query as an ordered list of layer-wise subqueries that measure the distribution of values across nodes in that layer.
A layer's subquery can be a wildcard (\query{*}) that matches any distribution, an integer that defines the number of nodes in that layer with a non-zero value, or a probability distribution (list) of numbers in the range $[0,1]$ that defines the relative node scores in a layer.
We incorporate the modifiers not (\query{!}), greater than (\query{>}), and less than (\query{<}) to expand query expressivity.
A query matches an instance if every layer in its fitted abstraction graph matches its corresponding subquery.
We can use the results of a query to understand how frequently an alignment pattern occurs and what its common outcomes are.
For instance, given a human abstraction graph with three levels, we can query for instances where the model distributes its confidence over multiple leaf nodes in the same subgraph as \query{[*,~[1],~>1]} (\cref{fig:interface-interactivity}D).
In the interface, we provide informative pre-defined queries in natural language to help users get started with querying.

\paragraph{Concept Co-confusion}
The concept co-confusion component instantiates the \conceptcoconfusion metric (\cref{eq:concept-coconfusion}) to reveal pairs of concepts that the model commonly confuses (\cref{fig:interface}E).
It is visualized as a list of concept pairs, sorted by their \conceptcoconfusion.
Each concept in a pair is colored based on its level-1 concept and shown above a sparkline~\citep{tufte2006beautiful} representing the pair's \conceptcoconfusion.
Users can filter the list of concept pairs based on attributes of an individual concept (i.e., its height, depth, name, and if it is a label) or the concept pair (i.e., if they are connected, share a parent, or share an ancestor) (\cref{fig:interface-interactivity}E).
Selecting any pair updates the rest of the interface to show instances that contributed to that pair's \conceptcoconfusion.

\paragraph{Instance List}
Finally, to allow users to drill down into the model's alignment on individual decisions, the abstraction alignment interface displays a fitted abstraction graph (\cref{sec:method-model-outputs}) for every dataset instance (\cref{fig:interface}F).
Each fitted abstraction graph is displayed as a nested horizontal bar chart, where a bar represents a node in the fitted abstraction graph and its length represents that node's value.
Bars are colored based on the level-1 concept and the root bar shows a summary of the bars below it.
The fitted abstraction graphs are displayed next to the instance text or image and its true labels.
To visually indicate salient areas of the fitted abstraction graph, we bold the text corresponding to the instance's labels and any of its direct relatives.
When selections occur in other interface elements, the instance list updates to display the relevant instances.
We display summary statistics above the list, showing the number of instances selected and, in classification settings, the proportion of them that are correctly and incorrectly classified.

%% file: code/abstraction_alignment_propagation.tex
\lstset{
  language=Python,            %
  basicstyle=\ttfamily\footnotesize, %
  backgroundcolor=\color{lightgray}, 
  keywordstyle=\color{Thistle},      %
  commentstyle=\color{Periwinkle},     %
  stringstyle=\color{red},        %
  numberstyle=\tiny\color{codegray},
  numbers=left,                   %
  numberstyle=\tiny,              %
  stepnumber=1,                   %
  numbersep=5pt,                  %
  showspaces=false,               %
  showstringspaces=false,         %
  tabsize=2,                      %
  breaklines=true,                %
  breakatwhitespace=false,        %
  escapeinside={(*@}{@*)},        %
  captionpos=b,                   %
  framesep=0pt,
  frame=single,
  rulecolor=\color{lightgray},
  framextopmargin=3pt,
  framexleftmargin=3pt,
  framexbottommargin=3pt,
}

\begin{figure}[t]
\begin{lstlisting}
# Pseudocode to compute the model's fitted abstraction
# graph for one dataset instance.
def fit(abstractionGraph, model, outputs, instance):
  # Initialize the values in the human abstraction graph.
  fittedAbstractionGraph = abstractionGraph.copy()
  for node in fittedAbstractionGraph:
    node.value = 0

  # Set node values based on the model's confidence.
  probabilities = model(instance)
  for i, probability in enumerate(probabilities):
    node = fittedAbstractionGraph.getNode(outputs[i])
    node.value = probability

  # Propagate the values from leaf to root.
  for level in reverse(fittedAbstractionGraph.levels):
    for node in level:
      for child in node.children:
        node.value += child.value

  return fittedAbstractionGraph
\end{lstlisting}
\caption{
To compare model behavior with human abstractions, abstraction alignment computes a fitted abstraction graph for each model decision. 
First, we map the model's output space to concepts in the human abstraction graph.
Then, we assign each concept a value corresponding to the model's confidence in that concept or any of its descendants. 
The resulting fitted abstraction graph represents the model's confidence in a range of concepts across levels of abstraction.
}
\Description{Pseudocode to compute the model's fitted abstraction graph for one dataset instance.}
\label{lst:abstraction-fit}
\end{figure}

%% file: figures/interface_figure.tex
\begin{figure*}[t]
  \includegraphics[width=\textwidth]{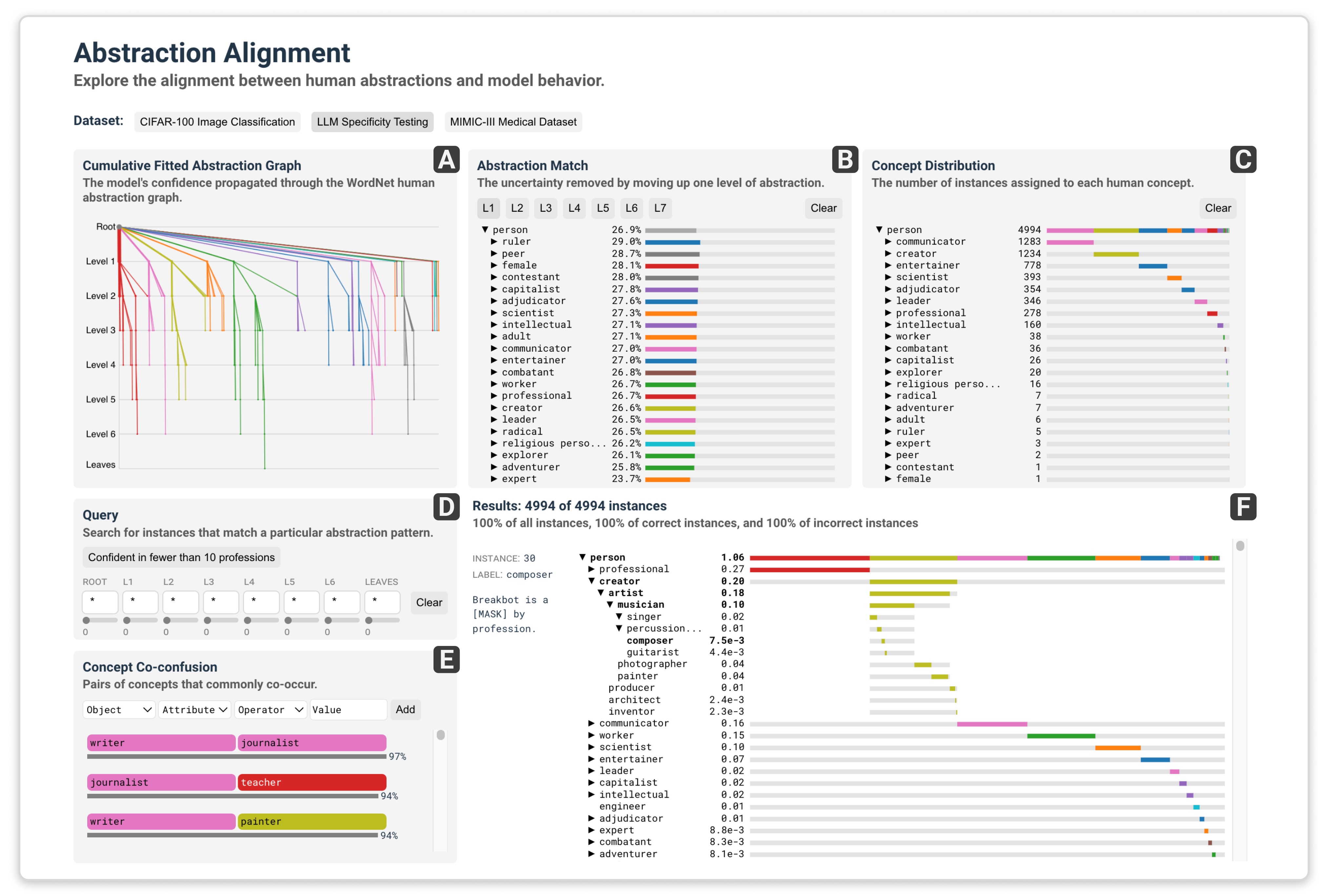}
  \caption{
  The abstraction alignment interface visualizes a model's alignment with human abstractions. It displays the cumulative fitted abstraction graph (A), aggregated \abstractionfit (B), concept distribution (C), and \conceptcoconfusion (E). Interacting with these panels or the query bar (D) updates the instance list (F) to show the fitted abstraction graphs of relevant inputs.
  }
  \Description[Screenshot of the abstraction alignment interface]{Screenshot of the abstraction alignment interface with size panels labeled A--F. The top left panel (A) is the Cumulative Fitted Abstraction Graph, a tree diagram with nodes and edges of varying colors representing the concepts and abstraction relationships. The top middle panel (B) is the Abstraction Match, a horizontal bar plot showing the abstraction match of top-level concepts. The top right panel (C) is the Concept Distribution, a horizontal bar plot showing the number of dataset instances within each concept. The middle left panel (D) is the Query view, showing a set of eight input boxes and sliders, one per level. The bottom left panel (E) is the Concept Co-confusion, showing pairs of concepts and a sparkline of their concept co-confusion score. The bottom right panel (F) is a list of dataset instances. One instance representing the input sentence "Breakbot is a [MASK] by profession" is shown next to a horizontal bar chart showing the model's confidence across abstraction graph concepts.}
  \label{fig:interface}
\end{figure*}

%% file: figures/interactivity_figure.tex
\begin{figure*}[!t]
  \includegraphics[width=\textwidth]{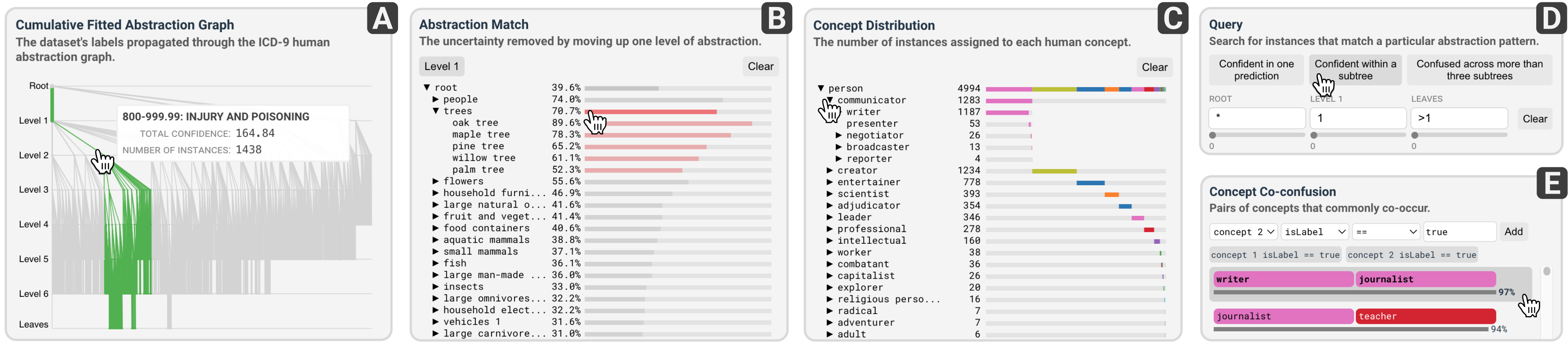}
  \caption{
  Interacting with the abstraction alignment interface allows users to explore alignment hypotheses.
  Users can select a concept (A–C), a concept pair (E), or define an alignment query (D) to update the interface with relevant dataset instances.
  }
  \Description[Screenshots of the abstraction alignment interface's interactivity]{Five panels show the interactivity of the abstraction alignment interface. From left-to-right, the cumulative fitted abstraction graph (A) is shown with the cursor hovering over a node in the graph corresponding to the concept "injury and poisoning". A tooltip displays the total confidence and number of instances corresponding to the concept. All nodes and edges not connected to the hovered node are greyed out. The abstraction match panel (B) is shown with a cursor hovering over the "trees" concept. The "trees" concept is in full color, while its children are at a lower opacity and all other concepts are greyed out. The concept distribution panel (C) is shown with a cursor expanding the concept "communicator" to show its direct children. In the query view (D), the cursor highlights a button "confident within a subtree" and the query below shows [*, 1, >1]. Finally, the concept co-confusion panel (E) shows the cursor hovering over a pair of clickable concept pairs.}
  \label{fig:interface-interactivity}
\end{figure*}

%% file: sections/04_case_studies.tex
\section{Evaluative Case Studies with Domain Experts}
\label{sec:case-studies}

By externalizing \revision{formal} human knowledge as an abstraction graph, abstraction alignment expands current alignment workflows from internalized comparison to iterative hypothesis testing. 
To evaluate this shift in perspective, we emulate three real-world alignment tasks across computer vision, natural language, and medicine.
First, in \cref{sec:interpreting-cifar-models}, we apply abstraction alignment to interpret an image classification model, finding that it expands interpretation from narrow questions about why a model made a specific decision to broad explorations of the human concepts it has learned.
Next, in \cref{sec:language-model-specificity}, we collaborate with researchers to benchmark the specificity of language model responses, revealing that abstraction alignment expands their conventional benchmarks of isolated pairwise comparisons to more comprehensive comparisons across the entire space of potential outcomes.
Finally, in \cref{sec:medical-dataset-analysis}, we leverage abstraction alignment earlier in the ML pipeline, using it with healthcare professionals to assess the human alignment of a medical dataset, revealing discrepancies between medical abstractions and their real-world usage.

\paragraph{Study Method}
To evaluate how abstraction alignment supports real-world alignment analysis, we collaborate with seven domain experts across two case studies: language model specificity (\cref{sec:language-model-specificity}) and medical dataset analysis (\cref{sec:medical-dataset-analysis}). 
We conducted in-depth, semi-structured interviews with each expert to assess how abstraction alignment influenced their analysis.
We began with questions about their domain expertise, alignment workflows, and desired outcomes, such as \textit{“Tell me about your role as a [title]?”} and \textit{“How do you currently measure alignment in [case study task]?”}.
Next, we introduced the abstraction alignment interface (\cref{sec:interface}) using tasks and datasets representative of their domain. 
Finally, we prompted experts to think aloud as they engaged with abstraction alignment to identify ways the model or dataset was aligned or misaligned with their domain knowledge.
This approach allowed us to understand experts' current processes, observe how abstraction alignment functioned in context, and assess its potential to address experts' alignment goals.
\revision{We discuss study limitations in \cref{sec:discussion}.}

\input{tables/participant_table}

We targeted expert participants to understand how abstraction alignment could impact real-world domains.
To identify experts, we reached out to authors of relevant literature, attendees of specialized conferences, and LinkedIn professionals with applicable expertise.
We purposively sampled~\citep{campbell2020purposive} seven participants, ensuring they had deep familiarity with their case study\,---\,language model participants regularly tested language models and medical dataset participants had extensive experience with medical codes (\cref{tab:participants}).

We conducted six video interviews each lasting 30--60 minutes (P6 and P7 opted to interview together as colleagues).
Our institution deemed our study exempt from full IRB approval, and participants received \$50 gift cards.
With consent, all interviews were recorded, resulting in 223 minutes of audio/video data and transcripts.
We conducted a thematic analysis, reviewing recordings and transcripts to code key observations, such as cases where participants recognized an alignment/misalignment (e.g., the dataset is missing domain-specific abstractions), expressed ways abstraction alignment facilitated/hindered their analysis (e.g., it replicated their existing experiment design), or identified an insight that led them to hypothesize about the downstream impact (e.g., the model is overly specific at the expense of correctness).
After analyzing recordings individually, we grouped codes into higher-level themes, and used them to structure the results in \cref{sec:language-model-specificity-qualitative} and \cref{sec:medical-dataset-analysis}.

\subsection{Interpreting Image Model Behavior}
\label{sec:interpreting-cifar-models}

\input{figures/cifar-case-study-figure}

A common interpretability task is understanding a model's mistakes; however not all mistakes are equally problematic.
For instance, in an autonomous driving task, mistaking a \concept{truck} for a \concept{bus} might be harmless, whereas models that mistake a \concept{truck} for the \concept{sky} have, unfortunately, caused real-world accidents~\citep{tesla2016tragic,lambert2016understanding}.
We are more likely to forgive the first mistake because it more closely aligns with our human abstractions\,---\,\concept{trucks} and \concept{buses} are both \concept{vehicles} and we treat them similarly while driving.
However, the latter mistake could indicate more severe model generalizability issues where the model's abstractions do not align with accepted human reasoning  (i.e., \concept{trucks} and the \concept{sky} are vastly different concepts that require different interactions).

Applying abstraction alignment to this setting, we find that it helps interpret model behavior and differentiate the severity of a model's mistakes by expanding the number and complexity of concepts we use to characterize model decisions.
In particular, we use abstraction alignment to interpret a ResNet20~\citep{he2016deep} computer vision model trained on CIFAR-100~\citep{krizhevsky2009learning}.
We use the CIFAR-100 class and superclass structure as the human abstraction graph, as it maps low-level classes, like \concept{truck}, into higher-level concepts, like \concept{vehicles}~\citep{krizhevsky2009learning}. 
The result is a human abstraction graph with 121 nodes across three levels of abstraction.
We compute each test image's fitted abstraction graph by applying a softmax to the model's outputs and propagating them through the human abstraction graph.

\subsubsection{Interpreting Model Decisions}
Analyzing the abstraction alignment of an individual instance can indicate why the model made a particular decision.
An instance's fitted abstraction graph represents how the model made its decision on that instance and the conceptual similarity of other options it considered.
For example, in \cref{fig:teaser}, we show three CIFAR-100 test images, their fitted abstraction graphs, and the model's output probability.
The probability distribution for each image follows an approximately $60/20/20$ split, so we might assume the model is similarly confused about each image.
However, the fitted abstraction graphs reveal that the model's abstraction alignment differs significantly across images.
In the top example, the model's probability is split between three classes within the same concept, indicating the model is confident the image is a \concept{tree} and simply unsure of the species.
Whereas, in the bottom instance, the model assigns probability to three distinct high-level concepts, including \concept{fruit and vegetables} and \concept{large omnivores and herbivores}, indicating that it is very confused about this image or has learned a relationship that does not align with human expectations (e.g., a color relation where \concept{elephants} and \concept{mushrooms} are both \concept{gray}).
In a real-world setting, we may be willing to overlook a model's abstraction-aligned errors (like in the top image) and may want to penalize the model for unaligned confusion even if its answer is correct (as in the bottom image).

Analyzing the abstraction alignment of an instance can also reveal places where human abstractions could be updated to better fit the modeling task. 
For instance, analyzing the fitted abstraction graph for the middle image in \cref{fig:teaser} shows that the model splits its output probability between three classes: \concept{shark}, \concept{whale}, and \concept{dolphin}.
Confusion across these three classes may seem abstraction aligned because they are \concept{large ocean animals}.
However, our human abstraction graph splits them into separate high-level concepts because it is based on biological properties, where \concept{sharks} are \concept{fish} and \concept{whales} and \concept{dolphins} are \concept{aquatic mammals}.
While the biological principles that separate \concept{fish} and \concept{aquatic mammals} (e.g., gills or blow holes) are important to zoology, they are visually subtle and unlikely to come across in low-resolution CIFAR-100 images.
Given the model only has access to images, it makes sense that it could learn a visual abstraction where \concept{sharks} and \concept{whales} are closely related.
If our use case requires the model to learn biological abstractions, we might consider training on a different dataset with images that visually distinguish biological properties or contain additional metadata about the animal.
However, if we are only interested in the model's visual alignment, we may define a new human abstraction graph based only on visual similarity.

\subsubsection{Uncovering Global Patterns in Model Behavior}
Analyzing abstraction alignment across many model decisions can identify recurring patterns of misalignment that can impact the model's generalizability.
To analyze abstraction alignment across an entire dataset, we measure how often a model exhibits a particular type of abstraction alignment/misalignment.
We define types of abstraction alignment as queries describing the number of nodes the model considers at each level and how it distributes its confidence across nodes.
To measure how often the model's decisions align with human abstractions we query for instances where the model is confident in a single class or becomes confident in a single high-level concept.
In \cref{fig:cifar-case-study}A, we see these instances represent nearly half of the model's decisions but only $18\%$ of model mistakes.
This means most of the model's mistakes are not harmless low-level errors, but the result of confusion at higher levels of abstraction.
Digging into this further, we query for instances where the model considers at least four disjoint concepts and find that almost a quarter of all instances and half of mistakes fall into this category.

\subsubsection{Identifying Conceptual Alignment}
Given the model and human abstractions seemingly conflict, we can use abstraction alignment to identify which human abstractions the model has learned.
To do so, we analyze the model's \abstractionfit view (\cref{fig:cifar-case-study}B) and find that the \concept{people} and \concept{tree} abstractions resolve a large proportion of the model's uncertainty, indicating that the model has learned those abstractions.
For instance, $74\%$ of the model's uncertainty on images of \concept{babies}, \concept{boys}, \concept{girls}, \concept{men}, and \concept{women} is resolved at the parent concept \concept{people}.
While our model only achieves $67.7\%$ test accuracy, seeing that it has learned some human abstractions may increase our trust that its confusion in these categories is harmless, particularly if our setting does not require fine-grained classifications.

Our abstraction alignment metrics also reveal areas where the model is misaligned with human abstractions.
For instance, the model's uncertainty is not accounted for by abstractions like \concept{vehicles~2} nor does it appear to learn animal categorizations like \concept{non-insect invertebrates} and \concept{medium mammals}.
In both cases, we might consider these results to be acceptable model performance in light of ill-fitting human abstractions. 
In particular, the CIFAR-100 hierarchy artificially restricts each high-level concept to contain exactly 5 children\,---\,a constraint that produces two nodes for \concept{vehicles} that arbitrarily distinguish their children rather than meaningfully capture abstracted patterns.
In contrast, although the animal categories are semantically meaningful, they reflect biological concepts like size (\concept{medium}) and  reproduction (\concept{mammals}) that are seemingly hard for a model to learn from 32x32 images.
If learning accurate biological abstractions are important for our task, then we may prefer to train on an alternate dataset that more precisely expresses these characteristics; on the other hand, if learning visual abstractions is acceptable, we may update our human abstraction graph to better reflect what can be learned from the data (i.e., categorizing animals based on visual similarity).
With abstraction alignment we have revealed the model's learned abstractions, cases of misalignment with human expectations, and mitigation strategies to improve model alignment and existing human abstractions.

\subsection{Benchmarking Language Model Specificity with ML Researchers}
\label{sec:language-model-specificity}

\input{figures/llm-case-study-figure}

An essential alignment task for generative language model researchers is ensuring models produce outputs at an appropriate level of abstraction~\citep{zheng2023does, kamalloo2023evaluating, zhang2023language, yoshikawa2023selective}.
For instance, given the input \inlinequote{What is Claude Monet's profession?}, we would prefer a model that gave a specific answer, like \concept{painter}, instead of an overly general answer, like \concept{worker}.
On the other hand, if the model is unsure of Monet's exact profession, then we'd prefer that it outputs a more general answer it is confident in, like \concept{artist}, than a specific but possibly incorrect guess, like \concept{photographer}.
Currently, researchers assess language model specificity using benchmark datasets containing input prompts paired with multiple correct outputs at different levels of abstraction~\citep{huang2023can, yona2024narrowing,si2021whats,min2020ambigqa}.
However, these benchmarks limit researchers to testing a small number of possible answers across a few levels of abstraction (typically 2--4).
This can result in an incomplete understanding of model accuracy since the dataset may inadvertently over penalize answers that humans consider synonymous but are not included in the labels.
Moreover, since they dichotomize between a set of correct answers and all other incorrect answers, they do not provide researchers with insight into how wrong a particular mistake is (e.g., \concept{photographer} is a better guess for Monet's profession than \concept{cowboy}).

In this case study, we evaluate abstraction alignment's ability to improve language model specificity testing through interactive analysis with experts (\cref{sec:language-model-specificity-qualitative}) and quantitative benchmarks (\cref{sec:language-model-specificity-quantitative}).
We apply abstraction alignment to measure the specificity of five BERT~\citep{delvin2019bert}, RoBERTa~\citep{liu2019roberta}, and GPT-2~\citep{radford2019language} language models.
We evaluate each model on the S-TEST~\citep{huang2023can} specificity benchmark dataset, containing sentence prompts for masked token prediction of the subject's occupation, location, and birthplace.
Each of the prompts is labeled with a corresponding specific answer and general answer.
For instance the prompt ``\textit{Lake Louise Ski Resort is located in [MASK]}'' is paired with the specific answer ``\textit{Alberta}'' and the general answer ``\textit{Canada}''~\citep{huang2023can}.
To create fitted abstraction graphs, we map the model's output distribution over every possible answer to words in a lexical graph.
We create the human abstraction graph by mapping the S-TEST specific answers to nodes in the WordNet DAG~\citep{miller1995wordnet, fellbaum1998wordnet}.
We compute edges between nodes using WordNet's hypernym/hyponym and holynym/meronym functions, creating an abstraction graph of precise and general answers related to the task.
Since WordNet is an extensive lexical graph, it contains many concepts relevant to occupations and locations, making it a valuable proxy for human lexical knowledge on these tasks.
For example, it expands occupation specificity analysis from two concepts at two levels of abstraction to over 1,500 concepts across 9 levels of abstraction.

\subsubsection{Interactively Analyzing Model Specificity with NLP Experts}
\label{sec:language-model-specificity-qualitative}
To study how abstraction alignment impacts researchers' perspectives on language model specificity, we collaborate with three language model experts (\cref{tab:participants}).
All three experts have substantial experience benchmarking and evaluating language models, with P1 and P2 specializing in specificity analysis.
For these experts, specificity testing is critical for developing a comprehensive understanding of model behavior.
By expanding from a single ground-truth answer to a range of acceptable answers, specificity testing provides experts with a more nuanced assessment of model accuracy. 
It helps them develop a mental model of the model's behavior and estimate how it would behave on future inputs.
Despite these benefits, experts acknowledge that specificity testing is currently limited to a small number of acceptable answers across a few predefined levels of abstraction, restricting their ability to thoroughly evaluate model behavior against the full range of correct answers and abstraction levels observed in real-world language use.

A key specificity alignment task for experts was analyzing the model's preferred level of abstraction, so they could flag models whose outputs were uselessly general or misleadingly specific.
Experts' current specificity benchmarks contain a set of correct answers at various levels of abstraction, allowing them to test how often the model prefers a specific answer over a more general one.
However, by propagating model confidence through the abstraction graph, abstraction alignment broadened experts' perspectives on specificity testing.
Analyzing the confidence distributions across all nine abstraction levels in the fitted abstraction graphs, users observed that small probabilities on specific answers often summed to substantial confidence in more general responses and there were many cases where they could \inlinequote{elicit a different prediction by aggregating probabilities on all these very specific [answers]} (P1).
For example, in \cref{fig:llm-specificity}A, experts' traditional specificity benchmark would have penalized the model for incorrectly predicting Push Button is a \concept{photographer} since it is not a synonym for the correct answer, \concept{composer}.
But, by propagating the probabilities, abstraction alignment revealed that the model was conceptually correct just non-specific\,---\,i.e., all of its probability was assigned to types of \concept{artists}.
This insight demonstrated that the model's understanding was more nuanced than a traditional specificity benchmark could capture.
For P3, viewing the results through the abstraction alignment lens was \inlinequote{a much stronger claim for both specificity and categorization than just [comparing] to a [few] words.}

Experts also found that abstraction alignment allowed them test a range of hypotheses about model specificity that are not possible with current benchmarks.
During their analysis, experts often generated questions about the model's alignment, such as whether there are dissimilar professions that the model thinks are highly related (P2).
While current specificity benchmarks only consider a set of synonyms against all other incorrect options, abstraction alignment supported users in testing these alignment hypotheses by measuring the conceptual distance between model outputs. 
For example, to test their hypothesis, P2 examined the \conceptcoconfusion of occupation pairs that did not share an ancestor (\cref{fig:llm-specificity}B).
While they found it acceptable that the model assigned probability to co-occurring occupations, like \concept{journalist} and \concept{photographer}, they were surprised to see high \conceptcoconfusion between rare pairs of professions, such as \concept{lawyer} and \concept{painter}.
Experts worried that \inlinequote{these co-occurrences must [be coming] from the data distribution} (P1), suggesting more serious issues with the underlying training data that could cause \inlinequote{models trained on the same data to have similar co-occurrences} (P1).
Relatedly, P3 identified that \concept{professional} was assigned some probability across all 4,994 instances (\cref{fig:llm-specificity}C), hypothesizing that it was due to overuse of the word \concept{professional} in the dataset outside of an occupation context, such as \inlinequote{[person] is a professional}.
Since these findings suggested dataset artifacts were impacting the model's alignment, P2 wanted to confirm that these correlations existed in the dataset, and if so, \inlinequote{inform people who create data [that they] should be careful about these co-occurrences, they cause hallucinations}.
As a result, experts found that abstraction alignment expanded the types of specificity tests they perform, from narrow questions about a model's accuracy to broad hypotheses about the model's human-alignment.

Finally, beyond analyzing model specificity, experts hypothesized that abstraction alignment could improve model generation.
As P1 described, even \inlinequote{when a model is not confident, it will still produce a very specific answer, but [you know its not confident because] when you sample multiple responses, you will get different answers.}
In response to this phenomena, some model generation methods improve accuracy by relaxing the requirement for specificity through repeatedly sampling the model's output and identifying consistent details across the results~\citep{yona2024narrowing}.
However, P1 hypothesized that abstraction alignment could be an alternative method for improving model accuracy.
Instead of selecting the model's most probable answer (which is likely overly specific and incorrect), they were interested in using the fitted abstraction graphs to select the most specific concept above a particular confidence threshold.
For example, in \cref{fig:llm-specificity}A, instead of generating the model's most likely answer, \concept{magician} (which is incorrect), with abstraction alignment, they could generate a higher-level and higher-confidence answer, \concept{artist} (which is correct).
This example highlights the versatility of abstraction alignment, showing that by viewing models through this lens, experts not only generated new hypotheses for specificity analysis but also uncovered novel strategies for model generation.

\input{tables/llm-specificity-table}

\subsubsection{Quantitatively Comparing Model Specificity}
\label{sec:language-model-specificity-quantitative}

While interactively exploring abstraction alignment expanded experts' qualitative analysis, it also improves traditional quantitative specificity benchmarks by generating a more diverse range of testable hypotheses. 
Existing specificity benchmarks are limited to comparing the model's probability across a small set of correct answers~\citep{huang2023can, yona2024narrowing,si2021whats,min2020ambigqa}. 
As a result, existing specificity benchmarks, like S-TEST, only measure the model's preference between one specific and one general answer~\citep{huang2023can}. 
Instead, by leveraging human lexical abstractions, abstraction alignment expands the number of possible answers and represents the relationships between answers. 
Thus, it enables us to test a broader range of specificity questions, such as how often the model prefers any specific answer to any general answer or whether it prefers a correct answer at any level of abstraction over an incorrect but task-specific answer.

To quantify specificity, we use \subgraphpreference (\cref{eq:subgraph-preference}) to compare the model's preference for answers in different regions of the lexical abstraction graph (\cref{tab:language-table}).
As a baseline, we recreate \citet{huang2023can}'s specificity metric by comparing the model's probabilities in the dataset-defined specific and general labels ($P(s, g)$).
Next, we expand this metric to test specificity across additional words and levels of abstraction.
Instead of testing one specific and one general answer, we compare all answers more specific than the specific label (specific label and its descendants) to all answers more general than the specific label (specific label's ancestors) ($P(s_{\downarrow}, s_{\uparrow})$).
Finally, we extend these metrics even further, testing whether the model prefers a correct answer at any level of abstraction to an incorrect but task-related answer by comparing all answers related to the specific label to all task-related words ($P(s_{\updownarrow}, t)$).

Benchmarking models with abstraction alignment reveals aspects of model behavior overlooked by prior metrics.
Existing metrics indicate that language models only have a slight preference for specific answers, with most $P(s, g)$ between $40-70\%$~\citep{huang2023can}.
However, by expanding to a larger set of possible answers, abstraction alignment reveals that language models actually have a strong preference for specific answers.
For example, \texttt{bert-large} prefers a specific answer on over $80\%$ of instances across all tasks.
This result suggests that by only comparing two answers, prior metrics are too strict and do not account for variety of model preferences, whereas abstraction alignment more accurately reflects model specificity by considering a larger set of human-aligned answers.

Beyond making specificity testing more accurate, abstraction alignment also allows us to test other aspects of specificity.
Using $P(s_{\updownarrow}, t)$, we can test the model's preference for a correct answer at any level of abstraction to an incorrect answer related to the task.
For instance, when predicting ``\textit{Enrico Castellani is a [MASK] by profession}'' we compare all answers that are direct ancestors or descendants of the specific label \concept{painter} to all other answers related to any other occupation in the dataset.
While previously we found models prefer a specific correct answer to a general correct answer, here, we find that models often prefer a incorrect answer to any correct answer.
This is not always correlated with accuracy or other specificity metrics\,---\,for instance, \texttt{gpt-2} has the lowest accuracy and specificity on occupation prediction but the highest preference for correctness.
By expanding the set of concepts we can analyze, abstraction alignment expands traditional benchmarks, exposing otherwise hidden aspects of model behavior.

\subsection{Analyzing Medical Dataset Encodings with Healthcare Professionals}
\label{sec:medical-dataset-analysis}

\input{figures/mimic-case-study-figure}

We also investigate how abstraction alignment can enhance participatory dataset analysis by identifying discrepancies between the abstractions users expect models to learn and those codified in the dataset.
Machine learning models build their internal representations by learning correlations between input features and output labels in their training data~\citep{paullada2021data}.
However, the correlations encoded in the dataset are not always the correlations we expect the models to learn~\citep{birhane2021large,caliskan2022gender} due to labeling inconsistencies~\citep{northcutt2021pervasive}, limited data diversity~\citep{de2023fair}, or societal biases~\citep{klein2024data, bolukbasi2016man, torralba2011unbiased}.
As a result, a core task for ML developers is to audit their datasets, ensuring they reflect task expectations~\citep{jones2023no, vaughn2020dataset, suresh2021framework}.
Yet, this process is often challenging, especially in applied settings where ML developers lack the domain expertise needed to deeply understand and evaluate the dataset's content~\citep{nazi2024large, bhattachary2023lessons,bhattacharya2024exmos}.
While participatory AI workflows address this gap by engaging domain experts in dataset curation and auditing~\citep{cooper2024from,birhane2022power,lam2022end}, eliciting high-quality feedback remains difficult due to the size and complexity of ML datasets~\citep{deng2023understanding}. 

To investigate how abstraction alignment can facilitate participatory data analysis, we collaborated with four healthcare professionals to compare the medical abstractions encoded in a clinical ML dataset against global health standards.
Specifically, we focus on a clinical coding task where medical coders label narrative descriptions of patient hospital stays with ICD-9 codes representing diseases the patient had (e.g., \concept{730: bone infection}) and procedures they received (e.g., \concept{78.0: bone graft})~\citep{dong2022automated}.
These codes, derived from the World Health Organization's (WHO) 9th revision of the International Classification of Disease (ICD-9) hierarchy~\citep{worldhealthorganization1978icd9}, are critical for justifying costs to insurance companies, tracking epidemiological data, and reporting morbidity statistics~\citep{Alexander2003Overview}.
AI-automated clinical coding is an active area of research~\citep{dong2022automated,edin2023automated}, with many models using the MIMIC-III dataset for training and evaluation~\citep{johnson2016mimica,johnson2016mimicb}.
Ideally, models should learn to apply ICD-9 codes based on the WHO's code guidelines, but discrepancies between real-world code application and prescribed guidance can arise due to coder inexperience, coding system complexity, and intentional misuse to increase insurance payout~\citep{OMalley2005Measuring}.
Since MIMIC-III contains real-world patient records, its code labels may deviate from the ICD-9 hierarchy's intended use.
This is a critical consideration for ML developers, as models trained on MIMIC-III risk learning and perpetuating these misaligned abstractions during deployment.

We simulate a participatory MIMIC-III dataset audit with four clinical coding experts\,---\,two professional medical coders (P4, P5) with decades of experience coding clinical notes at major hospitals and two ICD experts (P6, P7) at national health organizations (\cref{tab:participants}).
While these experts have limited ML knowledge, they possess deep knowledge of the medical coding task and ICD-9 guidelines, representing domain experts that participatory AI workflows aim to include.
Since the MIMIC-III dataset contains thousands of clinical notes and ICD-9 codes~\citep{johnson2016mimica,johnson2016mimicb}, it would be infeasible for experts to manually inspect the dataset's alignment by reading each clinical note and comparing the labeled codes to their expectations.
However, we hypothesize that abstraction alignment may help scaffold dataset alignment by enabling structured comparisons between the dataset’s code application and the ICD-9 hierarchy, which our experts study and use in their professions.

To apply abstraction alignment in this setting, we use the ICD-9 hierarchy~\citep{mullenbach2018explainable} as the human abstraction graph, containing 21,116 nodes over 7 levels of abstraction.
The nodes represent the ICD-9 codes and edges abstract from low-level codes (e.g., \concept{461.1: acute frontal sinusitis}) to higher-level code groups (e.g., \concept{460-466: acute respiratory infections}).
The ICD-9 graph is a relevant proxy for \revision{formal} human knowledge as it represents the collective health standards our medical coders were trained on during their education.
Since in this setting we are comparing the dataset's encodings to human abstractions, we use the dataset's labels to compute the fitted abstraction graphs.
Here, the fitted abstraction graphs reflect the relationships between the codes assigned to the same medical note, where a node's value is equal to the number of codes corresponding to that node or any of its descendants.
We apply this procedure to the MIMIC-III test set, resulting in $3,372$ fitted abstraction graphs.

To ensure the MIMIC-III dataset aligns with medical use, all four experts sought to confirm that the distribution of ICD-9 codes in the dataset reflected the disease and procedure frequencies they experienced in real-world hospital settings.
Since clinical notes often contain verbose codes representing highly-specific diseases and procedures, it is challenging to make sense of pattens in the raw label distribution.
However, abstraction alignment allowed experts to examine the code distribution at their preferred level of abstraction by aggregating specific code labels (e.g., \concept{427.31: atrial fibrillation}) into progressively higher-level code groupings (e.g., \concept{427: cardiac dysrhythmias} and \concept{390–459: diseases of the circulatory system}).
Using the instance's fitted abstraction graphs, experts began by inspecting the dataset’s distribution across the four top-level ICD-9 code groups (\concept{00–99.99: Procedures}, \concept{001–999.99: Diseases and Injuries}, \concept{V01–V86.99: Supplementary Health Factors}, and \concept{E800–E999.9: Supplementary Causes of Injury and Poisoning}) and iteratively drilled down into subcategories and individual codes.
For example, by inspecting the top-level color distribution of individual clinical notes (\cref{fig:mimic-case-study}B), P4 confirmed that the dataset's code distribution aligned with their expectations, \inlinequote{[This distribution] makes sense, because [on a single visit], you could code twenty diagnosis codes, two procedure codes, and maybe a V or E code.}

However, expanding the overall code distribution and moving down a level of abstraction to more closely inspect the disease codes caused P5 to worry that the distribution skewed towards \concept{cardiovascular} and \concept{endocrine} diseases, with very few codes related to \concept{pregnancy and childbirth} (\cref{fig:mimic-case-study}A).
Since the dataset was sourced from a large hospital, it made sense to P5 that the most commonly applied codes matched common diseases, like heart disease and diabetes.
However, since P6 and P7 are responsible for nationwide code usage, they were concerned that the dataset might not accurately represent code usage in specialty departments, like stand-alone OB-GYN clinics.
Models trained on this dataset would experience a substantial distribution shift between the large hospital codes they saw during training and the specialty codes they would need to assign in practice.
For our experts, this discrepancy signaled the need for additional data collection from specialist departments that could be used to fine-tune models for specific use cases (P5) and downstream model evaluations stratified by hospital type to assess performance across varied clinical settings (P6).

Another important alignment task for the medical experts was to assess whether the MIMIC-III dataset captured meaningful relationships between co-occurring medical conditions and procedures.
In real-world practice, specific diseases and treatments often co-occur due to established medical correlations.
To explore the dataset's alignment with clinical reality, experts analyzed the filtered concept co-confusion view to test hypotheses about code pairs at different levels of abstraction (\cref{fig:mimic-case-study}C), such as how frequently applied codes co-occur with codes from an unrelated subgraph.
Experts were not surprised to see that the dataset commonly contains labels for codes like \concept{240--279.99: Endocrine, Nutritional, and Metabolic Diseases and Immunity Disorders} and \concept{401: Essential hypertension} because \inlinequote{patients with diabetes end up with hypertension} (P4).
However, they were concerned to see frequent co-labeling of \concept{other} and \concept{unspecified} codes, like \concept{270--279.99: Other Metabolic and Immunity Diseases} and \concept{285: Other and unspecified anemias}.
ICD-9 often contains codes representing specific variants of a disease as well as an \concept{other} or \concept{unspecified} catchall code.
Occasionally, applying an \concept{unspecified} code is appropriate when there is no way for the hospital to know the specific disease; however, the medical coders explained that high-quality coding often required them to request additional information from the doctor so they could apply the most accurate codes.
\begin{quote}
    \inlinequote{[This many unspecified] codes is a no, no. Our job is to code to the highest level of specificity because the more information you have, the more likelihood the insurance company understands what's going on and the claim gets paid.}~--~P5
\end{quote}
Experts worried that the frequent occurrence of \concept{unspecified} code labels in the MIMIC-III dataset could cause models to over apply \concept{unspecified} codes when they are unwarranted or not learn distinctions between the diseases contained within the \concept{unspecified} code, leading to billing issues and statistical discrepancies.

Our medical experts' abstraction alignment analysis of MIMIC-III reveals discrepancies between how ICD-9 codes are applied in the dataset and human expectations for disease classification.
These misalignments suggest that even models that achieve high performance on the dataset may not align with medical standards and could perpetuate code misapplication, leading to inaccurate insurance billing and epidemiological statistics.
However, abstraction alignment enables experts to engage in participatory dataset auditing, identifying these issues before models are trained and using their insights to guide corrective actions, like re-coding problematic patient records or reweighting the dataset to balance the frequency of over-applied codes.
Notably, abstraction alignment also enabled medical experts to identify limitations in the ICD-9 abstractions themselves.
In fact, the overuse of \concept{other} and \concept{unspecified} codes that our experts found via abstraction alignment corresponds to real-world changes the WHO made during the transition from ICD-9 to ICD-10 to increase code specificity~\citep{cartwright2013icd, who2022icd10}. 
These findings suggest that beyond supporting participatory dataset analysis, abstraction alignment can also uncover opportunities to refine human-designed abstractions, ensuring they better support both clinical practice and machine learning workflows.

%% file: tables/participant_table.tex
\begin{table}
\centering
\caption{
We evaluate abstraction alignment through case studies with seven experts in language model specificity (\sref{sec:language-model-specificity}) and medical dataset analysis (\sref{sec:medical-dataset-analysis}). 
}
\label{tab:participants}
\resizebox{\linewidth}{!}{%
\begin{tabular}{rlll}
\multicolumn{4}{l}{\textbf{Language Model Specificity Case Study (\sref{sec:language-model-specificity})}} \\
\addlinespace[2pt]
{ID} & {Title} & {Affiliation} & {Role} \\
\midrule
P1 &  Professor & University & Tests model specificity \\
P2 &  Research Scientist & Tech Company & Tests model specificity \\
P3 &  Project Manager & Tech Company & Builds LLM benchmarks\\
\addlinespace[14pt]
\multicolumn{4}{l}{\textbf{Medical Dataset Analysis Case Study (\sref{sec:medical-dataset-analysis})}} \\
\addlinespace[2pt]
{ID} & {Title} & {Affiliation} & {Role} \\
\midrule
P4 & Medical Coder & Medical Center & Codes clinical notes\\
P5 & Medical Coder & Medical Center & Codes clinical notes \\
P6 & ICD Manager & Health Org. & Oversees ICD usage\\
P7 & ICD Manager & Health Org. & Oversees ICD usage\\
\end{tabular}
}
\end{table}

%% file: figures/cifar-case-study-figure.tex
\begin{figure*}
  \includegraphics[width=\textwidth]{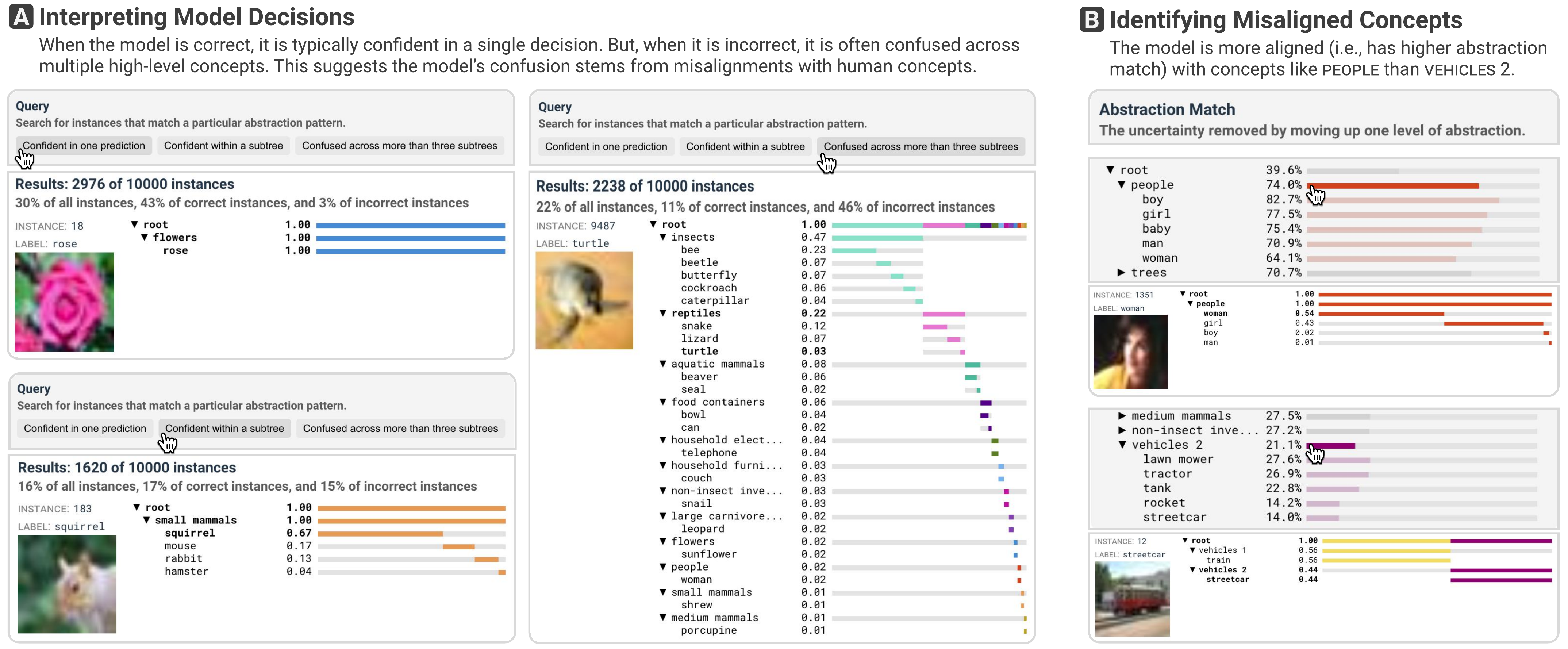}
  \caption{
  Abstraction alignment offers insights into the behavior of an image classification model. Querying alignment patterns distinguishes benign lack-of-specificity errors from more problematic misalignments (A), and analyzing \abstractionfit identifies which human concepts the model aligns with, highlighting potential failure cases (B).
  }
  \Description[Two panels depicting the case study on interpreting computer vision models.]{Two panels depicting the case study on interpreting computer vision models. Part A, "Identifying the Relationship Between Model Errors and Alignment" shows three queries and their fitted abstraction results. Part B, "Discovering Misaligned Concepts" shows two selections from the Abstraction Match widget, one for "people" and one for "vehicle" along with an example of their fitted abstraction results.}
  \label{fig:cifar-case-study}
\end{figure*}

%% file: figures/llm-case-study-figure.tex
\begin{figure*}
  \includegraphics[width=\textwidth]{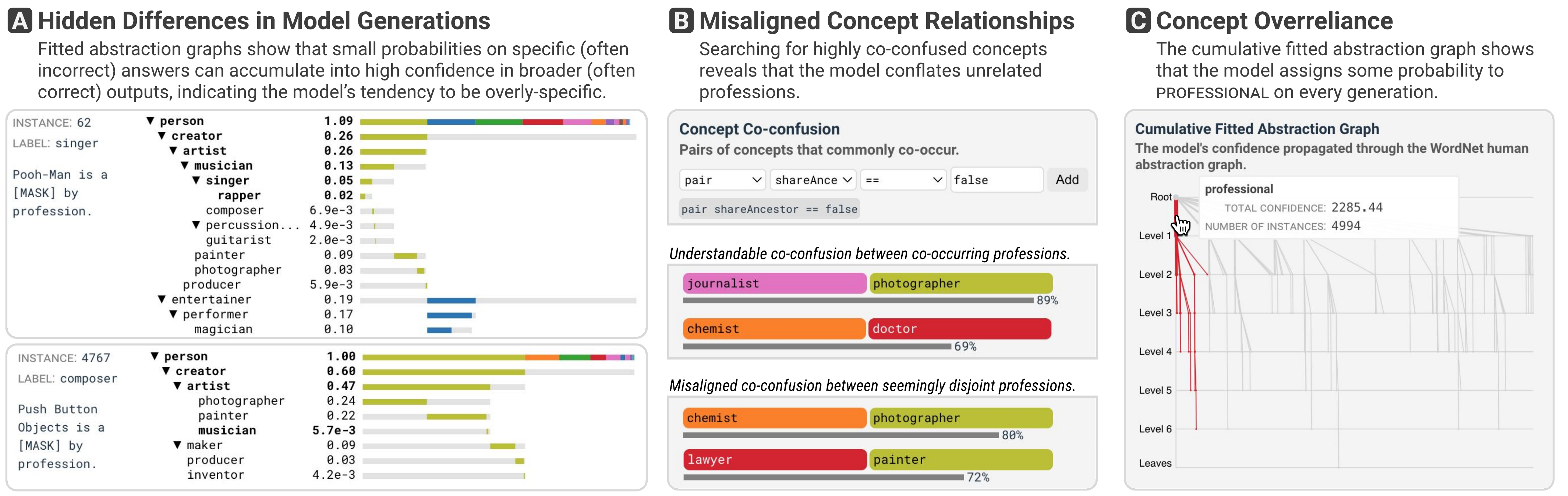}
  \caption{
  Abstraction alignment helps ML researchers better understand the specificity of generative language models, revealing the model's tendency to generate specific outputs at the expense of correctness (A), confuse seemingly unrelated concepts (B), and overrely on one particular concept (C).
  }
  \Description{Three panels depicting the case study on analyzing the specificity of generative language models. Part A "Hidden Differences in Model Confidence" shows two fitted abstraction graphs. Part B, "Misaligned Concept Relationships" shows the concept co-confusion panel filtered to pairs that do not share an ancestor. It shows two groups of concept pairs, "understandable co-confusion between co-occurring professions" and "misaligned co-confusion between seemingly disjoint professions". Part C "Concept Overreliance" shows the cumulative fitted abstraction graph with the cursor hovering on the node "professional". Its tooltip shows its total confidence is 2285.44 and its number of instances as 4994.}
  \label{fig:llm-specificity}
\end{figure*}

%% file: tables/llm-specificity-table.tex
\begin{table*}[t]
\centering
\caption{Abstraction alignment expands existing language model specificity benchmarks. We compute the \subgraphpreference and accuracy at 10 (A@10) of BERT~\citep{delvin2019bert}, RoBERTa~\citep{liu2019roberta}, and GPT-2~\citep{radford2019language} models on the S-TEST dataset's occupation, location, and birthplace tasks~\citep{huang2023can}. We compare existing metrics that test model preference between a specific and general answer ($P(s, g)$)~\citep{huang2023can} to abstraction alignment metrics measuring the model's preference for any specific answer to any general answer ($P(s_\downarrow, s_\uparrow)$) and a correct answer at any level of abstraction to an incorrect answer on the same task ($P(s_\updownarrow, t)$).}
\resizebox{\linewidth}{!}{%
\begin{tabular}{lrrrrrrrrrrrr} 
\multicolumn{1}{c}{} & \multicolumn{4}{c}{\textbf{Occupation}} & \multicolumn{4}{c}{\textbf{Location}} & \multicolumn{4}{c}{\textbf{Birthplace}} \\
\cmidrule(lr){2-5}\cmidrule(lr){6-9}\cmidrule(lr){10-13}
\textbf{Model} & \textbf{A@10} & $\pmb{P(s, g)}$ & $\pmb{P({s_\downarrow}, {s_\uparrow}})$ & $\pmb{P({s_\updownarrow}, {t}})$ & \textbf{A@10} & $\pmb{P(s, g)}$ & $\pmb{P({s_\downarrow}, {s_\uparrow}})$ & $\pmb{P({s_\updownarrow}, {t}})$ & \textbf{A@10} & $\pmb{P(s, g)}$ & $\pmb{P(s_{\downarrow}, s_{\uparrow}})$ & $\pmb{P(s_{\updownarrow}, t})$ \\ \midrule
\texttt{bert-base}     & 0.2844 & 0.7046 & 0.7901 & 0.0068 & 0.4316 & 0.4909 & 0.9591 & 0.2304 & 0.4142 & 0.6068 & 0.9992 & 0.2450 \\
\texttt{bert-large}   & 0.2214 & 0.7176 & 0.8240 & 0.0116 & 0.4564 & 0.4236 & 0.9704 & 0.2744 & 0.4214 & 0.5652 & 0.9967 & 0.2592 \\
\texttt{roberta-base}  & 0.2450 & 0.6180 & 0.7897 & 0.0751 & 0.3659 & 0.4999 & 0.9759 & 0.1790 & 0.2897 & 0.5448 & 1.000 & 0.2042 \\
\texttt{roberta-large} & 0.2244 & 0.7144 & 0.8238 & 0.0797 & 0.3905 & 0.4328 & 0.9785 & 0.2242 & 0.2321 & 0.4216 & 0.9989 & 0.2248\\
\texttt{gpt-2} & 0.1610 & 0.5728 & 0.5192 & 0.1684 & 0.1702 & 0.4825 & 0.4489 & 0.1348 & 0.3327 & 0.5972 & 0.9845 & 0.1959 \\
\end{tabular}%
}
\label{tab:language-table}
\end{table*}

%% file: figures/mimic-case-study-figure.tex
\begin{figure*}
  \includegraphics[width=\textwidth]{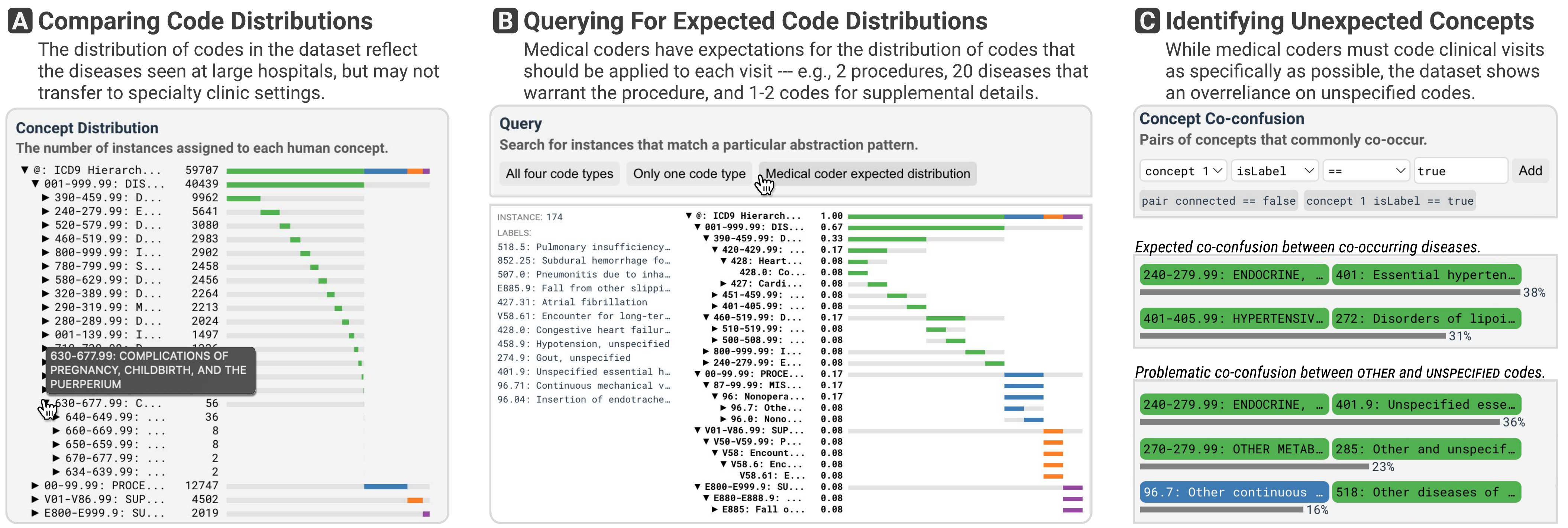}
  \caption{Abstraction alignment can also measure dataset alignment. Here, medical experts analyze a dataset of clinical notes and assigned codes. By comparing the distribution of medical code concepts (A \& B) and analyzing commonly co-coded concepts (C), abstraction alignment reveals discrepancies between dataset codes and real-world coding expectations.}
  \Description{Three panels depicting the medical dataset analysis case study. Panel A "Comparing Code Distributions" shows the concept distribution. The "diseases and injuries" concept is expanded and its child "complications of pregnancy, childbirth, and the puerperium" is hovered on. Panel B "Querying for Expected Code Distributions" shows the query "Medical coder expected distribution" and a fitted abstraction graph. Panel C "Identifying Unexpected Concepts" shows the concept co-confusion panel filtered to pairs of concepts that are not connected and where one concept is a label. Below, five pairs of concepts are shown grouped into "expected co-confusion between co-occurring diseases" and "problematic co-confusion between other and unspecified codes".}
  \label{fig:mimic-case-study}
\end{figure*}

%% file: sections/05_discussion.tex
\section{Discussion and Future Work}
\label{sec:discussion}

We consider abstraction alignment to be a methodology (i.e., an overarching strategy or conceptual foundation~\citep{crotty1998foundations}) for comparing the alignment of model-learned and human-encoded abstractions.
\revision{
In this paper, we have instantiated one method for measuring abstraction alignment that compares model outputs to human abstraction graphs.
This approach has proven valuable in assessing model and dataset alignment across computer vision, natural language, and medical domains.}
However, we expect there are many methods for measuring abstraction alignment and believe that developing these methods\,---\,tailored to different models, domains, and users\,---\,provides exciting opportunities for future work.

Our current method uses the model's output confidence as a proxy for its internal abstractions.
While this approach is model agnostic, allowing us to apply it across a range of models and flexibly extend it to dataset analysis, its reliance on model uncertainty limits our ability to measure abstraction alignment when models are confidently correct.
Alternative abstraction alignment methods could overcome this limitation by extracting abstractions directly from the model's internal representations, drawing inspiration from methods that identify a model's internal state~\citep{li2021implicit, hewitt2019structural} or its procedure for producing an output~\citep{olah2020zoom, elhage2021mathematical, wang2022interpretability, hanna2024does}.
New metrics could measure the alignment between these extracted model representations and existing human abstractions, revealing how a model's abstractions change across layers and evolve during training.

Another limitation of our abstraction alignment method is its dependence on human abstraction graphs as proxies for human knowledge.
Currently, abstraction alignment is limited to domains where abstraction graphs exist, such as linguistics~\citep{miller1995wordnet,fellbaum1998wordnet} and healthcare~\citep{worldhealthorganization1978icd9, johnson2016mimica, johnson2016mimicb}.
In some domains, abstraction graphs may not perfectly capture task semantics.
For instance, while WordNet maps specific areas (e.g. \concept{Patagonia}) to more general locations (e.g., \concept{Chile}), it is not a comprehensive location database and omits many towns (e.g., \concept{Coyhaique})~\citep{miller1995wordnet,fellbaum1998wordnet}.
As a result, in a location prediction task, WordNet may not include every location the model outputs and abstraction alignment may not fully capture the model's abstractions.
However, we are encouraged by the extensive research on knowledge graph generation~\citep{ji2020survvey,hogan2021knowledge}.
Incorporating these methods to generate human abstraction graphs in new domains could enhance abstraction alignment's applicability and effectiveness.

Further, while human abstraction graphs represent collective domain knowledge, they do not capture the diversity of knowledge that exists across individuals~\citep{suresh2021beyond}, so being abstraction-aligned does not guarantee universal alignment.
For example, doctors develop distinct medical abstractions based on their medical training and clinical experiences~\citep{cai2019hello}.
Thus, there are many ways to be aligned\,---\,e.g., representing a governing body's medical standards, hospital's practices, or individual clinician's perspective.
While, currently, abstraction alignment is limited to comparing against collective and \revision{formal} human knowledge, it offers an opportunity to envision personalized abstraction alignment methods, such those that codify individual knowledge or replace formal abstraction graphs with interactively specified representations.
Personalized abstraction alignment could improve human-AI collaboration by ensuring that the human and model are reasoning with the same abstractions.
Or, more interestingly, it could identify a user's optimal collaborator (e.g., a model with complementary abstractions) that expands the user's expertise and provides alternative perspectives.

Moreover, while the abstraction graph's DAG structure is computationally valuable, it imposes a rigid view of human knowledge that may not perfectly reflect human cognition.
Currently, the abstraction graph is based on Aristotelian concept theory where concepts define precise and discrete membership conditions~\citep{rosch2011slow}.
Thus, our abstractions are unambiguous\,---\,\concept{fruits} are seed-containing structures that develop from a plant's flower, so \concept{apples} and \concept{tomatoes} are both \concept{fruits}.
However, cognitive psychology has better represented human reasoning using graded concept theory, where membership is continuous~\citep{rosch1978principles}.
In this paradigm, a common fruit like \concept{apple} is a strong member of a \concept{fruit} whereas \concept{tomato} is a weak member because, while technically still a \concept{fruit}, we perceive them differently from prototypically sweet fruits.
Adopting graded concept theory may suggest a continuous measurement of abstraction alignment where the abstraction between concepts is weighted based on membership strength.

Likewise, while abstraction graphs provide a structured framework for measuring alignment, they inherently constrain the type of knowledge that can be represented, raising ethical implications about what it means to be aligned and whose knowledge we are aligning to~\citep{suresh2021beyond}.
These graphs are well-suited to domains with explicit and formalized knowledge but struggle to accommodate \revision{tacit,} subjective, or contextual knowledge that is harder to generalize and abstract into discrete concepts\revision{~\citep{hulpus2020knowledge,pan2024esparql,polanyi1966tacit,anderson2013architecture}}.
By privileging knowledge that is easily formalized, abstraction graphs may amplify dominant representations of knowledge, potentially marginalizing alternative ways of knowing~\citep{suresh2021framework,winner2017artifacts,feenberg2008critical}.
This risks reinforcing existing power dynamics, encouraging the development of models that perpetuate dominant worldviews~\citep{gordon1980power}.
Future alignment research should critically examine and document the perspectives we align to~\citep{Gebru2021datasheets} and explore more informal abstraction representations that better represent diverse forms of knowledge.

\revision{
Nevertheless, our case studies demonstrate that the current instantiation of abstraction alignment helps domain experts interpret model and dataset alignment.
While there are two limitations with the design of our case studies\,---\,they follow a largely exploratory protocol and only engage seven participants\,---\,these limitations map to our goal of understanding how grounding alignment analysis in abstractions can aid expert analysts.
Thus, we did not conduct comparative evaluations (e.g., against alternative alignment methods), nor do our case studies help us gauge the the value of our interface design choices or the usefulness of abstraction alignment in broader contexts (e.g., with less-expert analysts).
Nonetheless, our study design allowed us to gather rich, open-ended feedback from experts and evaluate abstraction alignment in real-world tasks.
Future work could complement our findings through large-scale comparative studies that directly contrast abstraction alignment to alternative alignment techniques across diverse participants, tasks, and domains.
For instance, studies could extend our exploration of abstraction alignment in dataset analysis by applying it to a real-world participatory dataset audit and comparing users' speed and findings against traditional auditing methods.
Additionally, a comparative study examining the insights users derive from their models using abstraction alignment versus established interpretability methods (e.g., saliency~\citep{molnar2022interpretable} or probes~\citep{li2021implicit, li2022emergent, hewitt2019structural}) could help contextualize the value of abstraction alignment and reveal how it complements existing approaches.
}

%% file: sections/06_conclusion.tex
\section{\revision{Conclusion}}
\revision{In this paper, we study \textit{abstraction alignment}: the agreement between a model's learned behavior and established human abstractions.
While current alignment workflows require analysts to mentally compare identified model concepts against their internal representations, abstraction alignment provides structure to alignment analysis by externalizing \revision{formal} human knowledge as a set of concepts and their abstraction relationships.
As a result, across interpretability workflows on computer vision and natural language tasks, abstraction alignment identifies recurring model misalignments, helping real-world domain experts build a mental model of the model's future behavior.
Beyond interactive analysis, abstraction alignment extends the expressiveness of quantitative model benchmarks, revealing aspects of model behavior overlooked by prior metrics, including underestimating models' preference for specific answers.
Finally, in a medical dataset analysis task with clinical experts, abstraction alignment reveals differences between the abstractions we expect models to learn and those codified in the dataset, suggesting improvements to existing human abstractions.}

%% file: sections/appendix.tex
\newpage
\section{Appendix}
\label{sec:appendix}

\subsection{Case Study Setup}
\label{sec:appendix-experimental-details}

Here we describe our case study setup (\cref{sec:case-studies}).
We provide code (\codeURL) and an interactive interface (\interfaceURL) to explore the results.

\subsubsection{CIFAR-100}
\label{sec:appendix-experimental-details-interpret}

In \cref{sec:interpreting-cifar-models}, we use abstraction alignment to interpret a CIFAR-100 image classification model.
We train a PyTorch~\citep{paszke2019pytorch} ResNet20 model~\citep{he2016deep} on the CIFAR-100 training set~\citep{krizhevsky2009learning} for 200 epochs with a batch size of 128.
We apply random crop and horizontal flip augmentations to the images following \citet{he2016identity}. 
We use cross-entropy loss optimized via stochastic gradient descent and Nesterov momentum~\citep{sutskever2013importance} (momentum = $0.9$; weight decay = $5e-4$).
We use a learning rate of $0.1$ and reduce it at epoch 60, 120, and 160 using gamma of 0.2.
The trained model achieves $67.7\%$ accuracy on the CIFAR-100 test set.

To apply abstraction alignment we use the CIFAR-100 class/superclass structure~\citep{krizhevsky2009learning} to form the human abstraction graph.
The graph contains 121 nodes across 3 levels\,---\,100 class nodes, 20 superclass nodes, and a root node.
We create a fitted abstraction graph for every dataset instance in the CIFAR-100 test set.
To do so, we compute the model's output probability for each image by applying a softmax to the model's output logits across all 100 classes.
Nodes that correspond to a CIFAR-100 class are assigned a value equal to the model's output probability for that class.
All other nodes' values are the sum of their reachable leaves' values.
For instance, \concept{tulip}'s value is the model's output probability that the image is a tulip, whereas \concept{flower}'s value is the sum of the model's output probability for \concept{orchid}, \concept{rose}, \concept{tulip}, \concept{sunflower}, and \concept{poppy}.

\subsubsection{Language Models and WordNet}
\label{sec:appendix-experimental-details-benchmark}
In \cref{sec:language-model-specificity}, we apply abstraction alignment to benchmark the specificity of language models.
Following the benchmarking procedure in \citet{huang2023can}, we compare pretrained \texttt{bert-base}~\citep{delvin2019bert}, \texttt{bert-large}~\citep{delvin2019bert}, \texttt{roberta-base}~\citep{liu2019roberta}, \texttt{roberta-large}~\citep{liu2019roberta}, and \texttt{gpt-2}~\citep{radford2019language} models from the LAMA benchmark~\citep{petroni2020how, petroni2019language}.
We test each model on the occupation, location, and birthplace tasks from the S-TEST dataset~\citep{huang2023can}.
Each S-TEST dataset instance is a text input paired with one specific and one general label.
For each model, we compute its top-10 accuracy, measured as the proportion of instances where the specific label was in the model's top 10 predicted tokens.

To measure abstraction alignment, we create an human abstraction graph for each of the occupation, location, and birthplace tasks.
For a task, we map each of its specific labels to its corresponding node (i.e., synset) in WordNet~\citep{miller1995wordnet, fellbaum1998wordnet}.
We do this process by searching for the specific answer label in the NLTK WordNet corpus\footnote{\url{https://www.nltk.org/howto/wordnet.html}}.
If there are multiple WordNet nodes that hit for a given search, we select the most appropriate node by manually inspecting their WordNet definitions.
Then, we expand the graph by including all direct ancestors and descendants of any specific label nodes.
The result is a human abstraction graph containing all the vocabulary words related to any of the data instances' specific labels.

To qualitatively explore abstraction alignment in \cref{sec:language-model-specificity-qualitative}, we create fitted abstraction graphs for every model decision on the occupation prediction task.
First, we compute the model's output probability across every word in its vocabulary for every data instance.
Then, for each data instance, we assign the model's output probabilities to their corresponding nodes in the human abstraction graph.
Finally, we propagate the values, such that each node's value is the sum of its value and its children's values.

To quantitatively benchmark the models in \cref{sec:language-model-specificity-quantitative}, we use the fitted abstractions to compute three specificity metrics, using \subgraphpreference (\cref{eq:subgraph-preference}).
Since, in this case, the model outputs map to concepts at many different levels of abstraction, we do not propagate the values through the fitted abstraction.
Instead, we assign each node a value corresponding to the model's probability of outputting that word. 
First, we replicate the specificity testing metric from \citet{huang2023can} (originally called $p_r$).
We compute it as $P(s, g)$, where $s$ is the single-node graph containing the specific label and $g$ is the single-node graph containing the general label.
Next, we compute $P(s_{\downarrow}, s_{\uparrow})$ to compare all words more specific than the specific label $s_{\downarrow}$ (specific label and its descendants) to all words at a higher level of abstraction than the specific label $s_{\uparrow}$ (specific label's ancestors).
Finally, we compute $P(s_{\updownarrow}, t)$ to compare ancestors and descendants of the specific label $s_{\updownarrow}$ to any other word in the task DAG $t$.

\subsubsection{MIMIC-III Medical Dataset with ICD-9 Codes}
\label{sec:appendix-experimental-details-dataset}
In \cref{sec:medical-dataset-analysis}, we apply abstraction alignment to analyze the abstractions in the MIMIC-III dataset~\citep{johnson2016mimica,johnson2016mimicb}.
The dataset contains textual medical notes paired with a set of ICD-9 code labels.
We use the ICD-9 medical hierarchy as the human abstraction graph~\citep{worldhealthorganization1978icd9}.
We pair the dataset's ICD-9 code labels with their corresponding code in the ICD-9 abstraction graph.
In this task, non-leaf nodes are codable\,---\,e.g., both \concept{282.6: sickle-cell anemia} and its direct parent \concept{282: hereditary hemolytic anemias} can be applied to the same medical note.
To compute the fitted abstractions graphs, we set the code node's value equal to one if the code was labeled on that instance and zero otherwise.
Then we propagate scores following \cref{lst:abstraction-fit}, setting each node's value equal to its value plus the summed values of its children. 
As a result the value of a node is equivalent to the number of times it or one of its children was labeled on the medical note.

\subsection{Compute Resources and Efficiency}
\label{sec:appendix-compute}
Time to compute abstraction alignment depends on the model, dataset, and human abstraction graph.
Extracting the model outputs/dataset labels, setting up the human abstraction graph, creating the fitted abstraction graphs, and measuring the abstraction alignment metrics take on the order of 10 minutes for the image model case study (\cref{sec:interpreting-cifar-models}), 5 minutes for the language model case study (\cref{sec:language-model-specificity}), and 30 minutes for the medical dataset analysis case study (\cref{sec:medical-dataset-analysis}).
We train and evaluate our models on 1 NVIDIA V100 GPU with 1TB of memory.

\subsection{Interface Implementation Details}
The abstraction alignment interface (\interfaceURL) uses Svelte to build responsive visualizations and the HTML5 Canvas to render performance-intensive charts.